\documentclass{llncs}
\usepackage{graphicx}
\usepackage{amsmath,amssymb} 
\usepackage{color}

\usepackage{subfigure}
\usepackage{array}
\usepackage{multirow}

\graphicspath{{imgs/}}

\newcolumntype{V}{>{$\vcenter\bgroup\hbox\bgroup}c<{\egroup\egroup$}}

\begin{document}

\pagestyle{headings}

\mainmatter

\title{Contextually Learnt Detection of Unusual Motion-Based Behaviour in Crowded Public Spaces}

\author{Ognjen Arandjelovi\'c}
\institute{
  Trinity College\\
  Cambridge, CB2 1TQ\\
  \texttt{ognjen.arandjelovic@gmail.com}}

\maketitle

\begin{abstract}
In this paper we are interested in analyzing behaviour in crowded
public places at the level of \emph{holistic motion}. Our aim is to
learn, without user input, strong scene priors or labelled data, the
scope of ``normal behaviour'' for a particular scene and thus alert
to novelty in unseen footage. The first contribution is a low-level
motion model based on what we term \emph{tracklet primitives}, which
are scene-specific elementary motions. We propose a clustering-based
algorithm for tracklet estimation from local approximations to
tracks of appearance features. This is followed by two methods for
motion novelty inference from tracklet primitives: (a) we describe
an approach based on a non-hierarchial ensemble of Markov chains as
a means of capturing behavioural characteristics at different
scales, and (b) a more flexible alternative which exhibits a higher
generalizing power by accounting for constraints introduced by
intentionality and goal-oriented planning of human motion in a
particular scene. Evaluated on a 2h long video of a busy city
marketplace, both algorithms are shown to be successful at inferring
unusual behaviour, the latter model achieving better performance for
novelties at a larger spatial scale.
\end{abstract}

\section{Introduction}
In recent years the question of security in public spaces has been attracting an increasing volume of attention. While the use of surveillance equipment has steadily expanded with it so has the range of problems associated with the way vast amounts of collected data are used. The inspection of video recordings by humans is a laborious and slow process, and as a result most surveillance footage is used not preventatively but rather \textit{post hoc}. Research on automating this process by means of computer vision aided inference has the potential to be of great public benefit and radically change how surveillance is conducted.


Public spaces such as squares and shopping centres are unpredictable and challenging environments for computer vision based inference. Not only are they rich in features, texture and motion, but they also continuously exhibit variation of both high and low frequencies in time: shopping windows change as stores open and close, shadows cast by buildings and other landmarks move, delivery lorries get parked intermittently etc. The focus of primary interest, humans, also undergo extreme appearance changes. Their scale in the image is variable and usually small, full or partial mutual occlusions and occlusions by other objects in the scene are frequent, with further appearance variability due to articulation, viewpoint and illumination.

As evidenced by a substantial corpus of published work, much computer vision work on the
modelling and understanding of human behaviour focuses on the recognition of articulated actions
\cite{YilkShah2008,DaviTyag2006,FarhTabr2008,TranSoro2008,HuanWangTanMayb2009,VeerSrinRoyCChel2009,LianShihShihLiao+2009}.
Although the analysis of crowds -- including their detection \cite{Aran2008}, tracking
\cite{AliShah2008} and counting \cite{ChanLianVasc2008} -- has started to develop some research momentum,
behaviour analysis in crowds on a more global level of holistic motion has
not been adequately addressed yet. Most previous work in this area considers scenarios far simpler
than we do in this paper --  camera viewpoint is often chosen so as to minimize occlusion of
objects of interests \cite{BrosCipo2006}
and only such specific types of scenes are considered which greatly restrict the variability of possible
motion patterns \cite{NguyPhunVenkBui2005,BranKett2000}. An often considered problem
is that of novelty detection in traffic \cite{ZhanHuanTanWang2007,HuXieTanMayb2004,HuXiaoFuXie+2009},
where motion constraints are much tighter. Not only does this greatly reduce
the space of possible motions over which learning is performed but it also makes it possible to
see in training all possible ``normal'' motion patterns, largely turning the problem to that of
achieving robustness to noise (usually through some form of clustering).


\section{Low-level building blocks}
The approach we describe in this paper can be broadly described as bottom-up. We first extract
trajectories of apparent motion in the image plane. From these a vocabulary of
elementary motions is built, which are then used to canonize all observed tracks.
Inference is performed on tracks expressed in this fixed vocabulary of motion primitives.
In this section we address low-level problems related to motion extraction and
its filtering, and the learning of motion primitives vocabulary.

\subsection{Motion Extraction}
As the foundation for inference at higher levels of abstraction, the extraction of motion
in a scene is a challenging task and a potential bottleneck. The key difficulty stems from the
need to capture motion at different scales, thus creating the compromise between reliability
and permanence of tracking. Generally speaking, the problem can be approached by
employing either holistic appearance, or local appearance in the form of local features. Holistic,
template based methods capture a greater amount of appearance and geometry, which can be
advantageous in preventing tracking failure. On the other hand, in the presence of frequent
full and partial occlusions, these methods are difficult to auto-initialize and struggle
with the problem of gradual bias drift as the tracked object's appearance changes due to
articulation, variable background and viewpoint, etc. All of these difficulties are very
much pronounced in the scenario we consider, motivating the use of local features.

Focusing on computationally efficient approaches, we explored several methods for detecting
interest points. Recently proposed Rosten-Drummond fast corner features \cite{RostPortDrum2008}
and Lowe's popular scale-space maxima \cite{Lowe2004} were found to be unsuitable due to lack of
permanence of their features: for an acceptable total number of features per frame, features that were
detected at some point in time remained undetected in more than 80\% of frames. Success was
achieved by adopting a simple method of tracking small appearance windows, in a manner similar
to Lucas and Kanade \cite{LucaKana1981}. Image region corresponding to the window $\mathcal{W}$
in frame $F_i$, is localized in subsequent frame $F_{i+1}$ by finding the translation which
minimizes the observed error between the two regions:
\begin{align}
  d_i^* = \text{arg}\min_d \int \int_{\mathcal{W}} \left[ F_{i+1}(x + \frac{d}{2}, y + \frac{d}{2}) - F_i(x - \frac{d}{2},y - \frac{d}{2}) \right]^2dx
  \label{e:klt}
\end{align}
This criterion is similar to that originally formulated by Lucas and
Kanade, with the difference that the expression is symmetrized in
time (i.e.\ with respect to $F_i$ and $F_{i+1}$). Further robustness
in comparison to the original method is also gained by performing
iterative optimization of \eqref{e:klt} in a multiscale fashion,
whereby $d_i^*$ is first estimated using smoothed windows and then
refined by progressively less smoothing. The best features (windows)
to track were selected as those corresponding to the $2 \times 2$
gradient matrices with the largest magnitude eigenvalues, as
proposed by Shi and Tomasi \cite{ShiToma1994}.

\subsubsection{Trajectory filtering.}
\begin{figure}[t]
  \centering
  \includegraphics[width=0.47\textwidth]{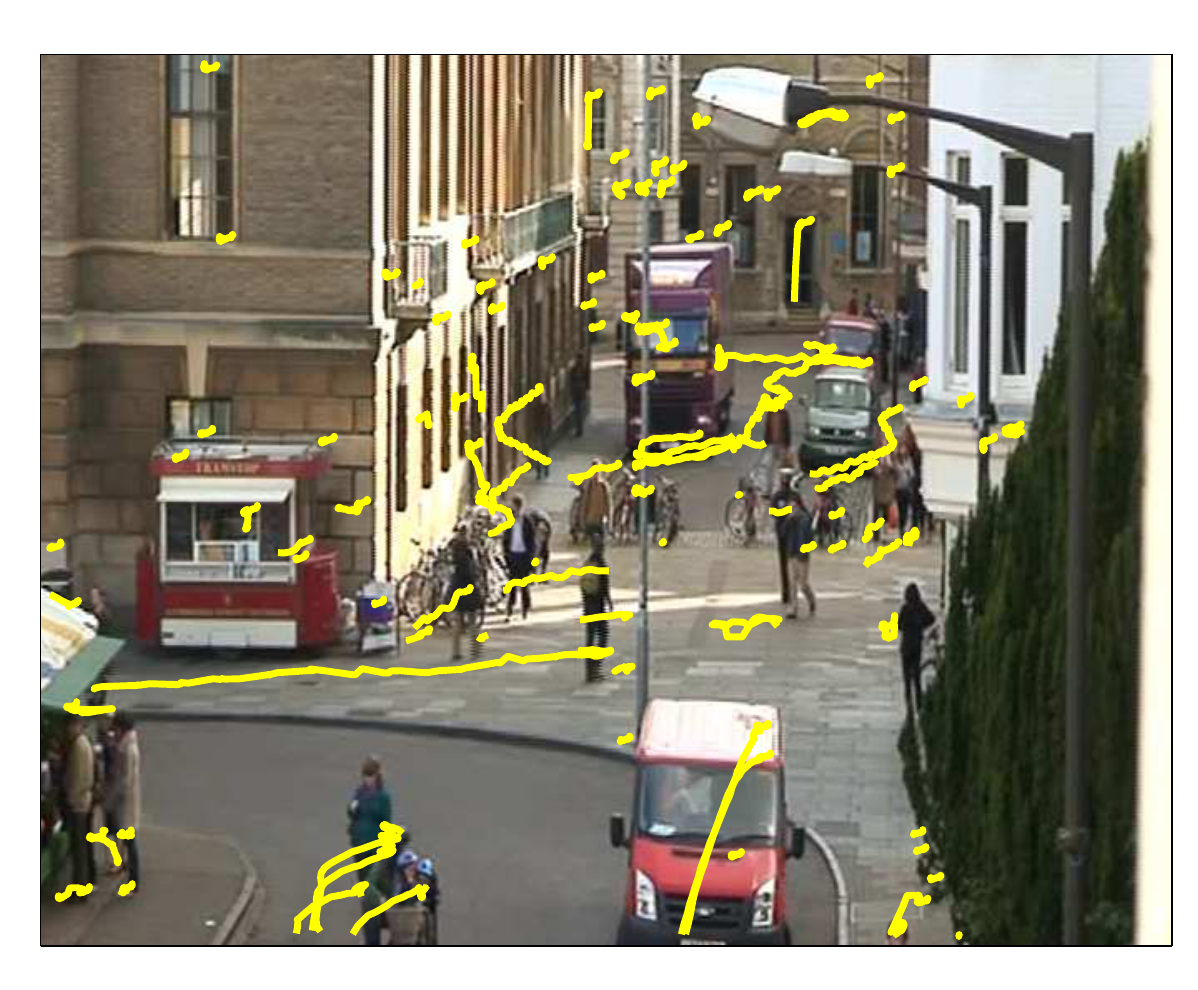}
  \includegraphics[width=0.47\textwidth]{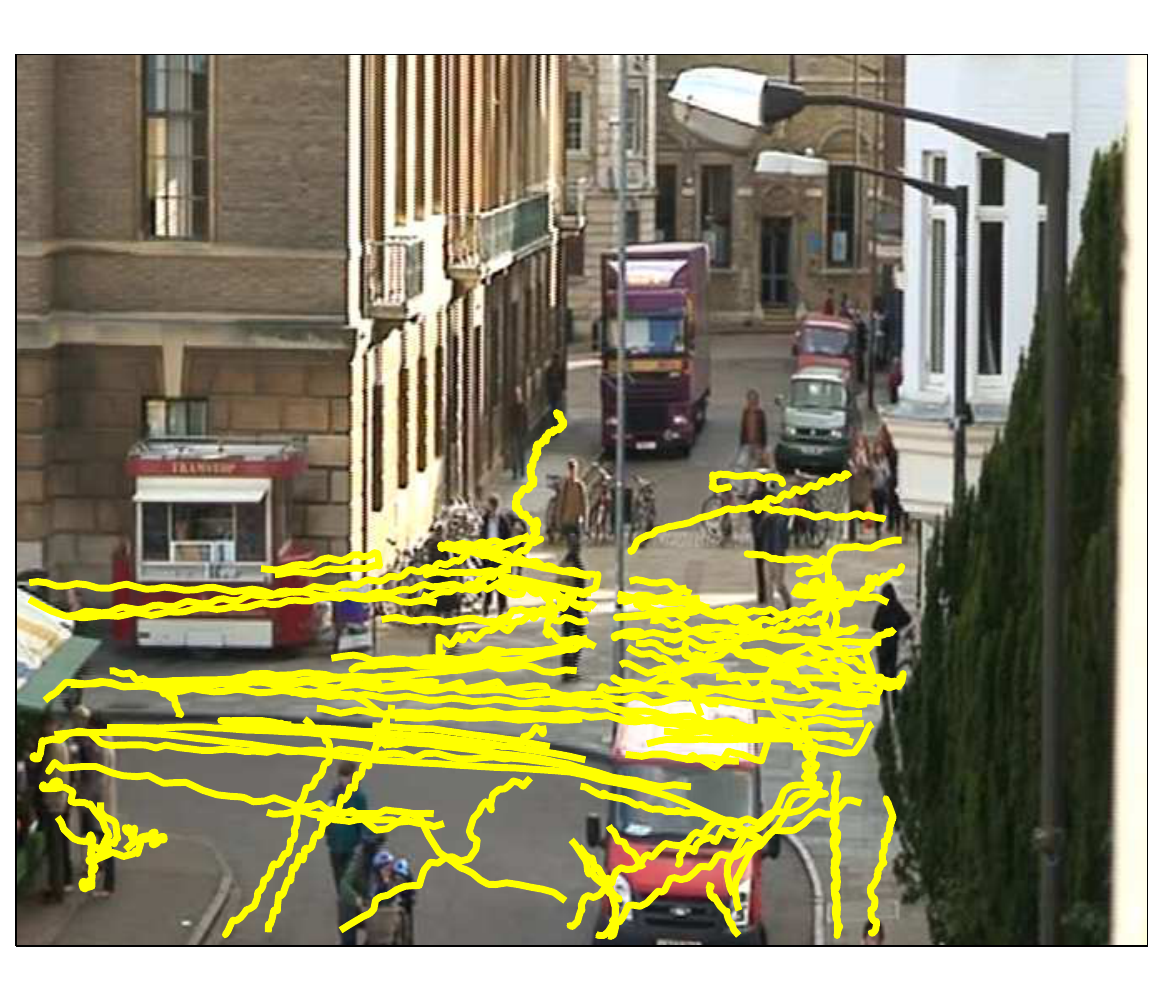}
  \caption{ (a) The first 100 feature tracks extracted by our feature tracker. Notice that there
            are many \emph{temporally} long tracks with minimal spatial extent due to the
            apparent motion of stationary features caused by small deflections of the camera by
            the wind. Filtering using the minimal displacement constraint can be used to reliably
            detect and discard such tracks. (b) The first 100 feature tracks after filtering.  }
  \label{f:tracks}
\end{figure}

Following the basic extraction of motion trajectories, we filter out uninformative tracks.
These are tracks which are too short (either due to low feature permanence or due to occlusion
of the tracked region) or which correspond to stationary features (possibly
exhibiting small apparent motion in the image plane due to camera disturbances, such as due
to wind), as illustrated in Figure~\ref{f:tracks}. We accept a track $\left\{(x_1,y_1),\ldots,(x_N,y_N)\right\}$
if $N \geq 30$ (i.e.\ it lasts for at least 30 frames, or 1.2s at 25fps) and:
\begin{align}
  \frac{1}{N-1}\sum_{i=1}^N \left[(x_i - \bar{x})^2 + (y_i - \bar{y})^2 \right] \geq \sigma_{min}^2,
\end{align}
where $\bar{x} = \sum_{i=1}^N x_i / N$ and $\bar{y} = \sum_{i=1}^N y_i / N$.

\subsection{Tracklet motion primitives}\label{ss:primitives}
People's motion trajectories in a scene can exhibit a wide range variability. However, not all of
it is relevant to the problem we address. For example, motion of interest is corrupted by noise and
at a short scale modulated by articulation. To reduce the effects of confounding variables on observed
motion, we represent all tracks using the same vocabulary of elementary motion primitives, inferred
from data, as illustrated conceptually in Figure~\ref{f:primitives}~(a).

\subsubsection{Inferring primitives.} We construct the vocabulary of motion primitives by clustering
\emph{tracklets} -- local, linear approximations of tracks. We extract a set of tracklets $t_i$ from a feature track
$T = \left\{(x_1,y_1),\ldots,(x_N,y_N)\right\}$ by first dividing the track into overlapping segments
$\left\{(x_{s(i)},y_{s(i)}),\ldots,(x_{e(i)},y_{e(i)})\right\}$ such that:
\begin{align}
  \Delta D = \sum_{j=s(i)}^{e(i)-1} \left\| \left(
                                         \begin{array}{c}
                                           x_j \\
                                           y_j \\
                                         \end{array}
                                       \right) -
                                       \left(
                                         \begin{array}{c}
                                           x_{j+1} \\
                                           y_{j+1} \\
                                         \end{array}
                                       \right)
                                        \right\|,
\end{align}
where $\Delta D$ is the characteristic scale parameter of the corresponding tracklet model. The $i$-th
tracklet is then defined by a numerical triplet consisting of its location and orientation $t_i = (\hat{x}_i, \hat{y}_i, \theta_i)$,
where:
\begin{align}
  &\hat{x}_i = \frac{1}{e(i)-s(i)+1} \sum_{j=s(i)}^{e(i)} x_j\\
  &\hat{y}_i = \frac{1}{e(i)-s(i)+1} \sum_{j=s(i)}^{e(i)} y_j\\
  &\hat{\theta}_i = \tan^{-1}\frac{y_{e(i)} - y_{s(i)}} {x_{e(i)} - x_{s(i)}} \hspace{10pt} (\text{mod } \pi)
\end{align}

All tracklets extracted from training data tracks are clustered using an iterative algorithm. In
each iteration, a new cluster is initialized with a yet unclustered tracklet as the seed. The
cluster is refined further in a nested iteration whereby tracklets are added to the cluster under
the constraint of maximal spatial and directional distances, respectively $\Delta Q$ and
$\Delta \Theta$, from both the seed and the cluster
centre. Cluster centre is then set equal to the mean of the selected tracklets and the procedure
repeated until convergence.

\begin{figure}[t]
  \centering
\begin{tabular}{VV}
  \includegraphics[width=0.45\textwidth]{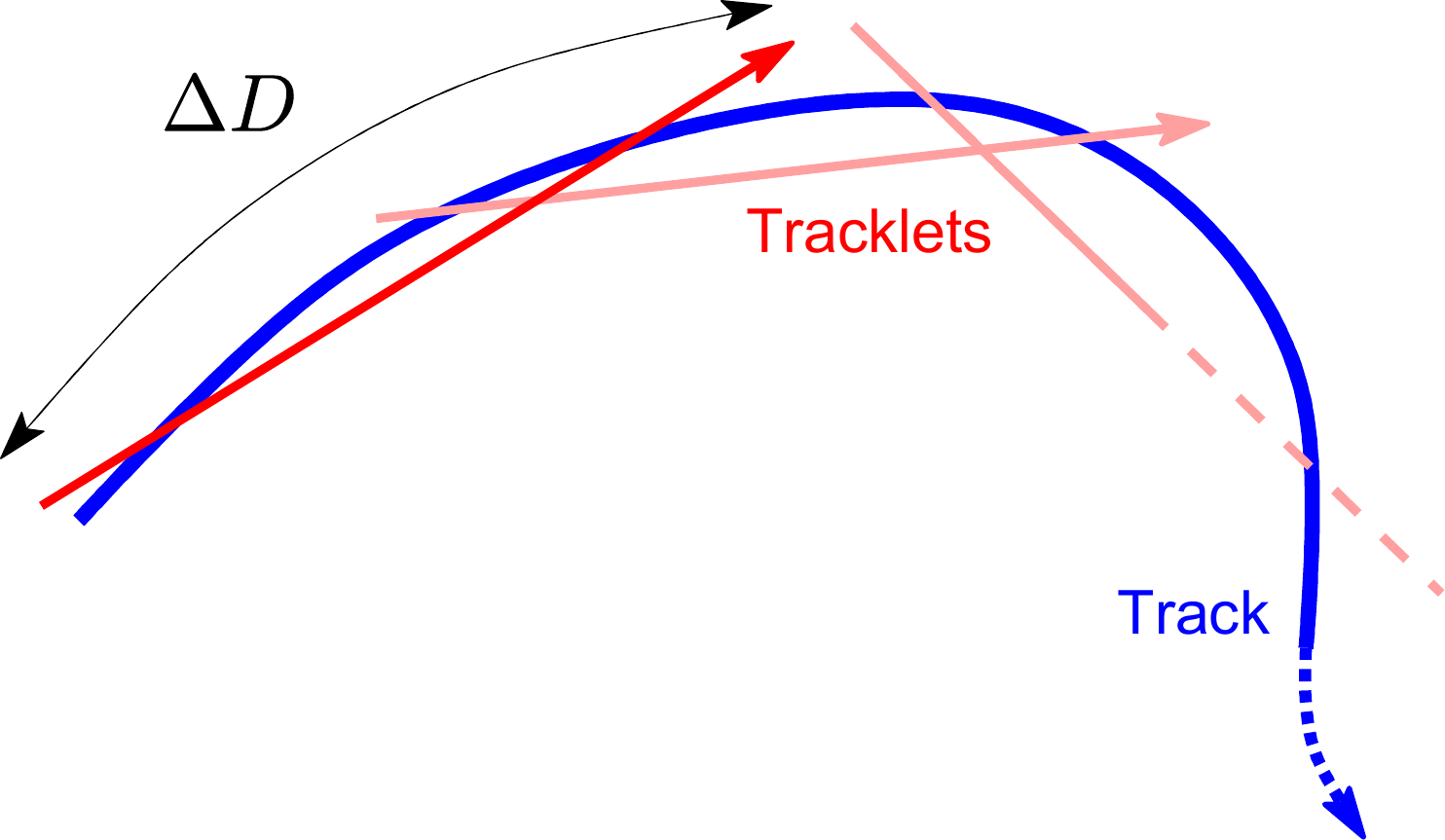} \hspace{25pt} &
  \includegraphics[width=0.37\textwidth]{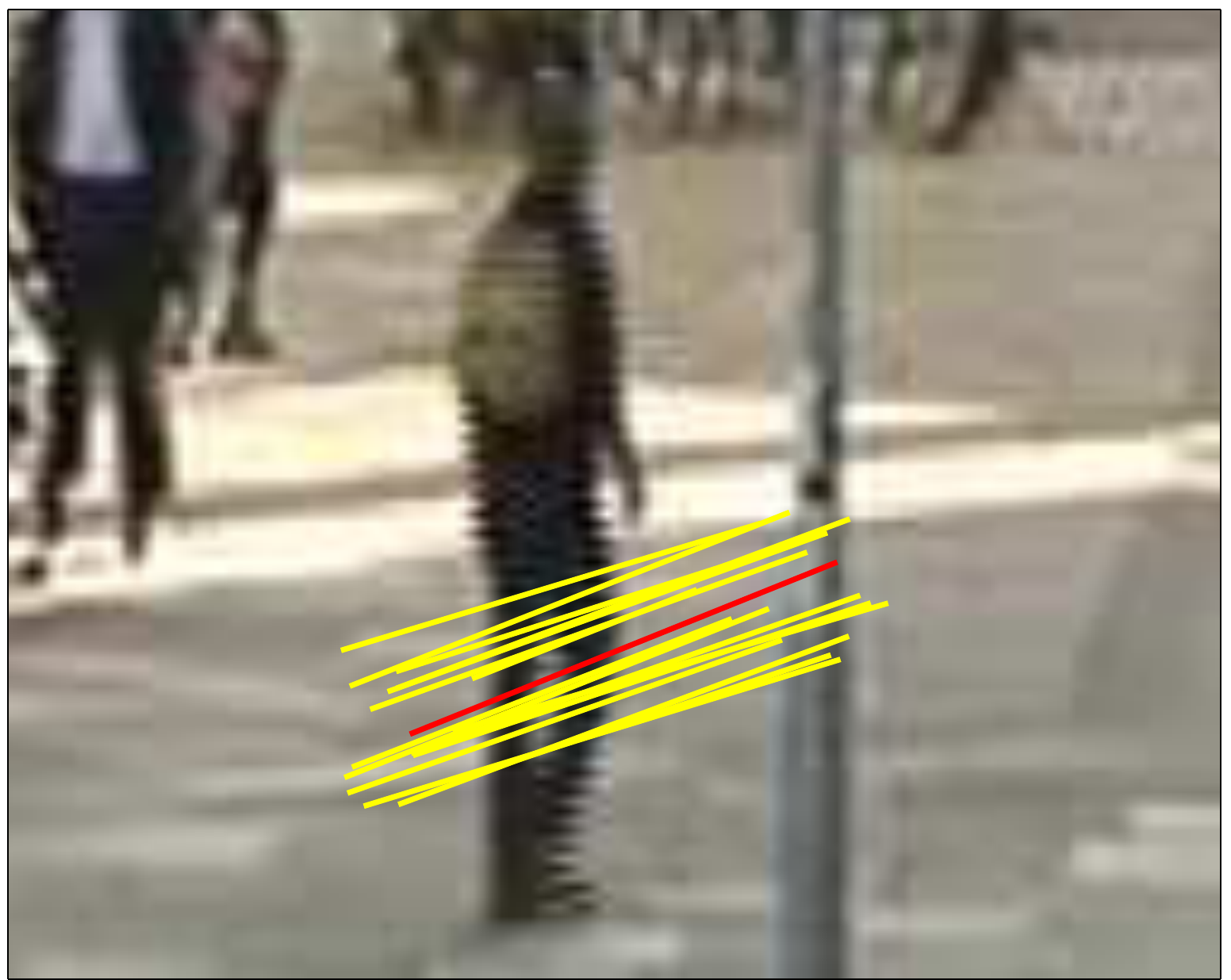}\\
  \vspace{-10pt}(a)\hspace{25pt} & \vspace{-10pt} (b) \\
  \end{tabular}
  \caption{ (a) By extracting a small set of motion primitives, all tracks can be expressed as
            sequences of primitives from the same vocabulary. (b) A motion primitive is
            estimated by clustering fragmented tracks -- \emph{tracklets} -- by spatial
            distance and directional agreement, and computing the mean (location and angle)
            of each cluster. }
  \label{f:primitives}
\end{figure}

Note that the equivalence of directions $\theta$ and $\theta \pm \pi$ introduces some difficulty
in the estimation of the cluster centre orientation. Specifically, it is \emph{not} appropriate
to average member directions using modulo $\pi$ arithmetic. First, note that the
problem is not always well posed, i.e.\ that it does not always have a unique solution.
Thus, we require that $\forall i.~\Delta \hat{\theta}_i < \pi/4$, where:
\begin{align}
    \Delta \hat{\theta}_i = \min\left \{(\hat{\theta}_i - \theta_c ) (\text{mod}~\pi), |\hat{\theta}_i - \theta_c| \right\}.
\end{align}
This condition ensures that the range of directions in a cluster is sufficiently constrained
that the mean direction is unambiguously definable. It is sufficient that $\Delta \Theta < \pi /4$
for this to be the case, which is certainly true in this paper, as a directional spread of over $\pi / 2$
within a cluster would produce meaningless tracklet groupings. Provided that a unique solution
exists, the following pseudo code summarizes the algorithm which correctly updates the cluster centre
direction $\theta_C$, given a list of directions $\hat{\theta}_1, \ldots, \hat{\theta}_N \in [0,pi)$ of the cluster
members:\\\\
\begin{tabular}{l}
\line(1,0){30}\\
  \hspace{8pt}$\Delta \theta = 0$ \\
  \hspace{8pt}for $i=1\ldots N$\\
  \hspace{18pt}$\Delta\hat{\theta}_i = \hat{\theta}_i - \theta_c$\\
  \hspace{18pt}if $(\Delta\hat{\theta}_i > +\pi/2)$ then $\Delta\hat{\theta}_i = \Delta\hat{\theta}_i - pi$\\
  \hspace{18pt}if $(\Delta\hat{\theta}_i \leq -\pi/2)$ then $\Delta\hat{\theta}_i = \Delta\hat{\theta}_i + pi$\\
  \hspace{18pt}$\Delta \theta = \Delta \theta + \hat{\theta}_i$\\
  \hspace{8pt}end\\
  \hspace{8pt}$\theta_c = \theta_c + \Delta \theta / N$\\
\line(1,0){30}\\
\end{tabular}\\\\\\
Figure~\ref{f:primitives}~(b) shows an example of tracklets grouped together and the corresponding
cluster centre which becomes a \emph{tracklet primitive}.


\subsubsection{Expressing tracks using primitives.}
A track is expressed in a particular tracklet model by diving it into overlapping segments,
computing the location and direction of each segment as before and, finally, associating
each segment $t_i = (\hat{x}_i, \hat{y}_i, \theta_i)$ with the most similar tracklet
$\mathfrak{T}_{j(i)} = (X_{j(i)}, Y_{j(i)}, \Theta_{j(i)})$:
\begin{align}
  t_i \longrightarrow
  \mathfrak{T}_{j(i)}:~~~j(i) = \text{arg} \min_k \epsilon(\mathfrak{T}_j, t_i),
\end{align}
where:
\begin{align}
  \epsilon(\mathfrak{T}_j, t_i) = (X_{j(i)} - \hat{x}_i)^2 + (Y_{j(i)} - \hat{y}_i)^2 +
  \left[ \frac{\Delta Q}{\tan(\Delta \Theta)} \tan(\Delta \hat{\theta}_i) \right]^2.
\end{align}
This functional form ensures that the angular contribution to distance is infinite for orthogonal
tracklets and approximately linear for small $\Delta \hat{\theta}$ (from Taylor's first order
expansion around zero: $\tan\alpha \approx \alpha$), whereas the proportionality factor ensures
that relative scaling of spatial and angular distances corresponds to the spread of cluster
tracklets.

\section{Two tracklet based motion models}
In the previous section we described our approach to extraction and
representation of motion in a scene. We now turn our attention to
the problem of learning a motion model from these low-level
representations and applying it to infer novelty in unseen data.

\subsection{First order Markov chain ensemble}\label{ss:ensemble}
The first model we introduce in this paper utilizes an ensemble of $K$ complementary
first order Markov chains
models to learn ``normal'' behaviour in a scene. The idea is that each model learns
behaviour on a different characteristic spatial scale. While this idea is now new,
it should be noted that our approach is different in that the
ensemble we construct is (in general) not a hierarchial one -- states describing behaviour on
longer scales do not consist of sequences of lower scale states. Rather, each model is built
independently by extracting tracklets and the corresponding motion primitives using different
characteristic scales, $\Delta D_1 < \ldots < \Delta D_K$, as proposed in Section~\ref{ss:primitives}.
The $k$-th model thus comprises learnt prior probabilities $P(\mathfrak{T}_i^{(k)})$ and
transition probabilities $P(\mathfrak{T}_j^{(k)} | \mathfrak{T}_i^{(k)})$ for tracklet
primitives at the corresponding scale.

Consider a particular novel feature track $T = \left\{(x_1,y_1),\ldots,(x_N,y_N)\right\}$.
In our model, the track is expressed independently in each of the $K$ vocabularies of
tracklet primitives:
\begin{align}
    T
 \longrightarrow
    \begin{cases}
       ~T_1 = \left\{ ~\mathfrak{T}_1^{(1)},~\ldots~,~\mathfrak{T}_{M_1}^{(1)}~\right\}\\
       \vspace{-10pt}\\
       ~T_2 = \left\{ ~\mathfrak{T}_1^{(2)},~\ldots~,~\mathfrak{T}_{M_2}^{(2)}~\right\}\\
       ~\hspace{35pt}\vdots \\
       ~T_K = \left\{ ~\mathfrak{T}_1^{(K)},~\ldots~,~\mathfrak{T}_{M_k}^{(K)}~\right\}\\
    \end{cases} \text{ where \hspace{5pt}} M_1 < M_2 < \ldots M_K,
\end{align}
where $T_1$ corresponds to the smallest scale of interest and $T_K$ the largest. Each of
the chains can then be used to compute the log-likelihood estimate corresponding
to its scale, which we average to normalize for differing track lengths:
\begin{align}
   R_k(T)= \frac{1}{M_k}
       \left[ \log P(\mathfrak{T}_1^{(k)}) + \sum_{i=2}^{M_k} \log P(~\mathfrak{T}_i^{(k)}~|~\mathfrak{T}_{i-1}^{(k)}~) \right].
    \label{e:logl}
\end{align}
Thus, the task of deciding if motion captured by $T$ sufficiently conforms to behaviour
seen in training is reduced to inference based on log-likelihoods $R_1(T),\linebreak[1]\ldots,\linebreak[1] R_K(T)$.
This would not be a difficult problem if both positive and negative training data (i.e.\ both unusual and normal motion patterns) was available, or if log-likelihoods corresponding to different models were directly comparable. However, a representative amount of positive training data is difficult to obtain in this case. Furthermore, although each $R_i(T)$ is normalized for track length in \eqref{e:logl}, the range of variation of its value is dependent on the model's characteristic scale. This is a consequence of lower entropy (generally) of larger scale models, which have fewer states (tracklet primitives).

To solve this problem, we transform the average log-likelihoods of all models to conform to
the same cumulative distribution function of the lowest scale (highest entropy) model:
\begin{align}
   R_k \longrightarrow \hat{R}_k = \mathcal{C}_1^{-1}\left[~\mathcal{C}_k(R_k)~\right]
\end{align}
where $\mathcal{C}_k(R)$ is the cumulative distribution function of the
average log-likelihood of the $k$-th Markov chain model, estimated from the training data set:
\begin{align}
   \mathcal{C}_k(R) = \int_{-\infty}^R p_k(r)~dr.
\end{align}
We then compute the conformance of the track to the overall
multiscale model as the minimal conformance to models over different
scales:
\begin{align}
   \rho_1(T) = \min_k \hat{R}_k
\end{align}

\subsection{Pursuit-constrained motion model}\label{ss:cmm}
In the previous section we described an approach to learning the range of normal motion in a
scene which treats a feature trajectory as a sequence of states corresponding to
extracted tracklet primitives. To make the parameter estimation practically tractable,
the sequence of states was modelled as a first order Markov chain which inherently restricts
the scope of the model to aggregating single-step behaviour. Progressively less spatially
constrained behavioural characteristics were captured by multiple tracklet models, each with
a different characteristic scale.

What this approach does not exploit is the structure of observed motion governed by the
of \emph{intentionality} of persons in the scene (whether they are on foot or using a vehicle).
While it is certainly the case that if unlimited data was available the described purely
statistical model would eventually learn this regularity, this insight can help
us achieve a higher degree of generalization from limited data which is available in practice.
Our idea is based on the simple observation that people perform motion with the aim of reaching
a particular goal and they generally plan their it so as to minimize the invested effort,
under the constraints of the scene (such as the locations of boulders and paved areas, or places
of interest such as shopping windows). Consequently, we concentrate on learning the distribution
of traversed trajectory lengths between two locations in a scene, rather than the exact paths taken
between them (a far greater range of possibilities).

Unlike in Section~\ref{ss:ensemble}, we now express a feature track $T$ as a sequence of tracklet
primitives \emph{only} in the vocabulary of the smallest scale of interest:
\begin{align}
    T \longrightarrow T_1 = \left\{ ~\mathfrak{T}_1^{(1)},~\ldots~,~\mathfrak{T}_{M_1}^{(1)}~\right\}.
\end{align}
For each pair of tracklets $\mathfrak{T}_i$ and $\mathfrak{T}_j$ ($i,j=1,\ldots,M_1$)
from the sequence we compute the corresponding distance $L_{ij}$ between them along the path. Since the
tracklet primitives were estimated using a single scale model with the characteristic scale parameter
$\Delta D_1$ (see Section~\ref{ss:primitives}), by construction this distance is given by:
\begin{align}
  L_{ij} =  (i-j)~\Delta D_1/2.
\end{align}

The track is thus decomposed into  $M_1(M_1-1)/2$ triplets $(\mathfrak{T}_i,\mathfrak{T}_j,L_{ij})$.
We wish to estimate $p(\mathfrak{T}_i,\mathfrak{T}_j,L_{ij})$. By expanding the joint probability as:
\begin{align}
  p(\mathfrak{T}_i,\mathfrak{T}_j,L_{ij}) = P(\mathfrak{T}_i)~~P(\mathfrak{T}_j~|~\mathfrak{T}_i)~~p(L_{ij}~|~\mathfrak{T}_i,\mathfrak{T}_j),
\end{align}
we can see that the first two terms -- the prior probability of the $i$-th primitive and the
probability of $i\rightarrow j$ transition -- can be learnt in a similar manner as for the Markov
chain based model described previously. On the other hand, the last term corresponding to the distribution of
possible path lengths between the $i$-th and $j$-th primitive, is computed by modelling it using a normal distribution:
\begin{align}
  p(\mathfrak{T}_j ~|~ \mathfrak{T}_i, L_{ij}) = \mathcal{N}( L~|~\bar{L}_{ij}; \sigma_{ij}).
\end{align}
We estimate its parameters -- the mean $\bar{L}_{ij}$ and standard deviation $\sigma_{ij}$ -- using
transitions between primitives observed in the training data set.

Finally, the conformance of a novel track to the learnt motion model is computed as the log of the lowest
probability tracklet primitive transition contained within the observed motion:
\begin{align}
  \rho_2(T) = \min_i \min_j \bigg[\log p(\mathfrak{T}_j, \mathfrak{T}_i, L_{ij})\bigg].
  \label{e:rho2}
\end{align}

\section{Evaluation}
\begin{figure}[t]
  \centering
  \subfigure[]{\hspace{3pt}\includegraphics[height=0.28\textwidth]{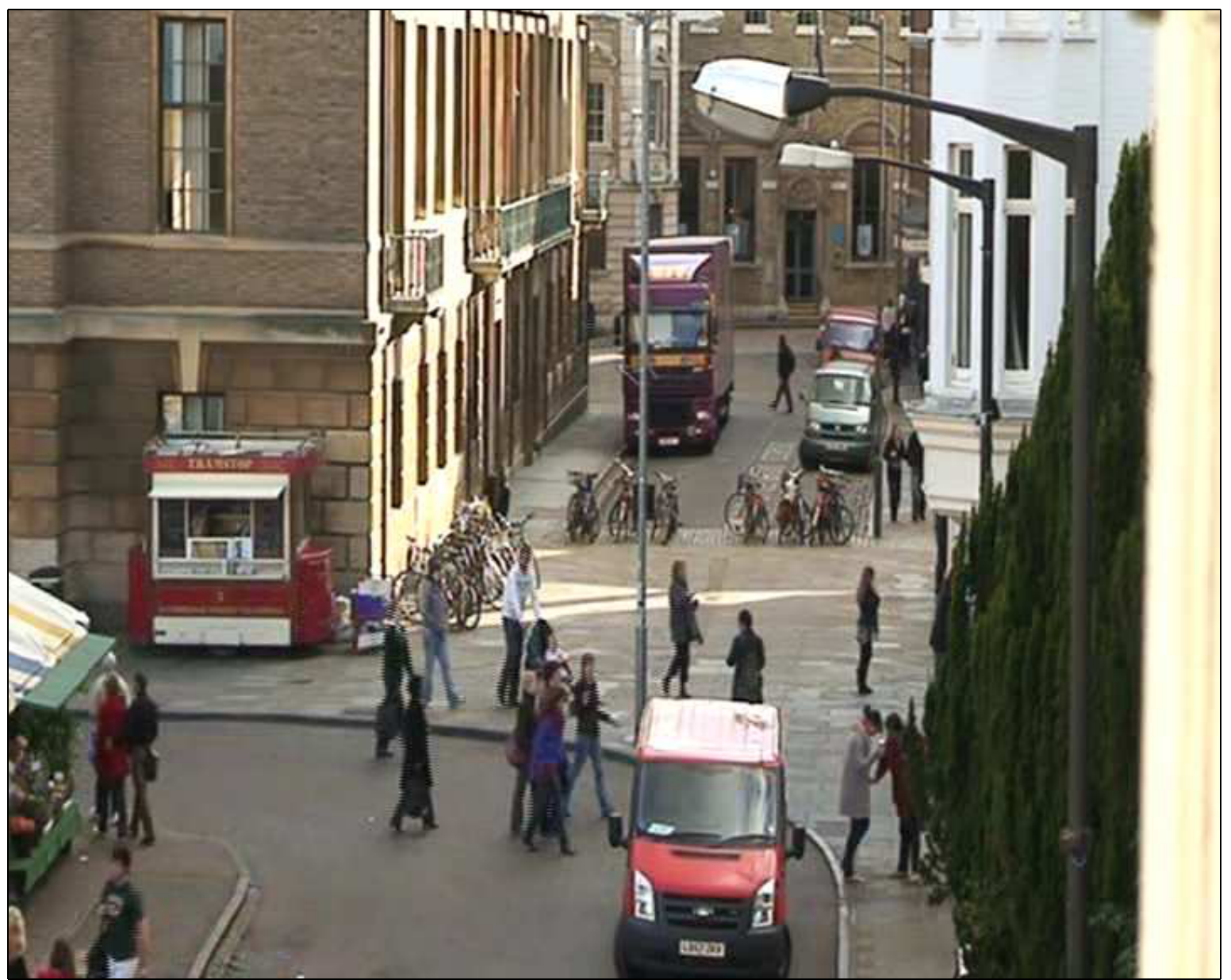} \hspace{3pt}}
  \subfigure[]{\hspace{3pt}\includegraphics[height=0.28\textwidth]{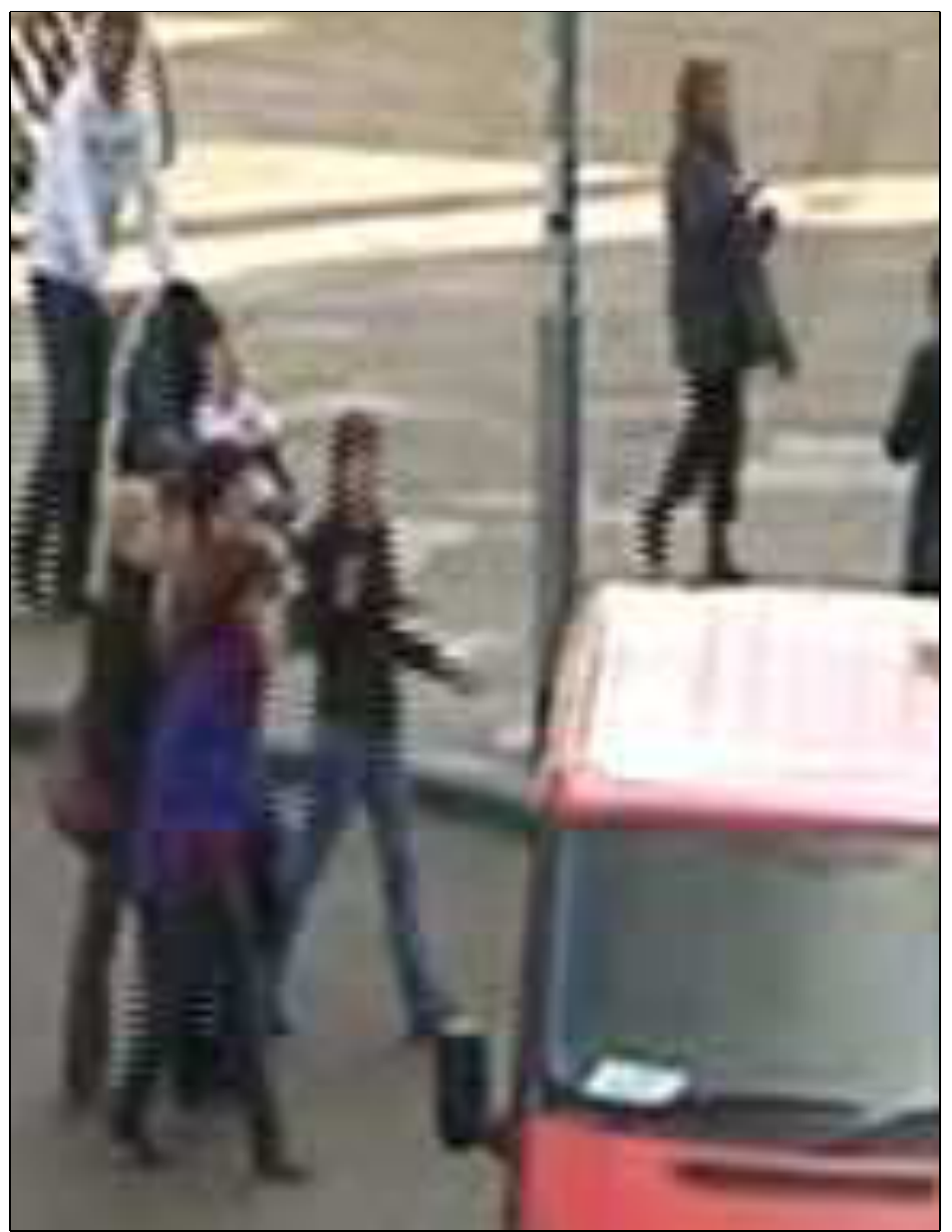}\hspace{3pt}}
  \subfigure[]{\hspace{3pt}\includegraphics[height=0.28\textwidth]{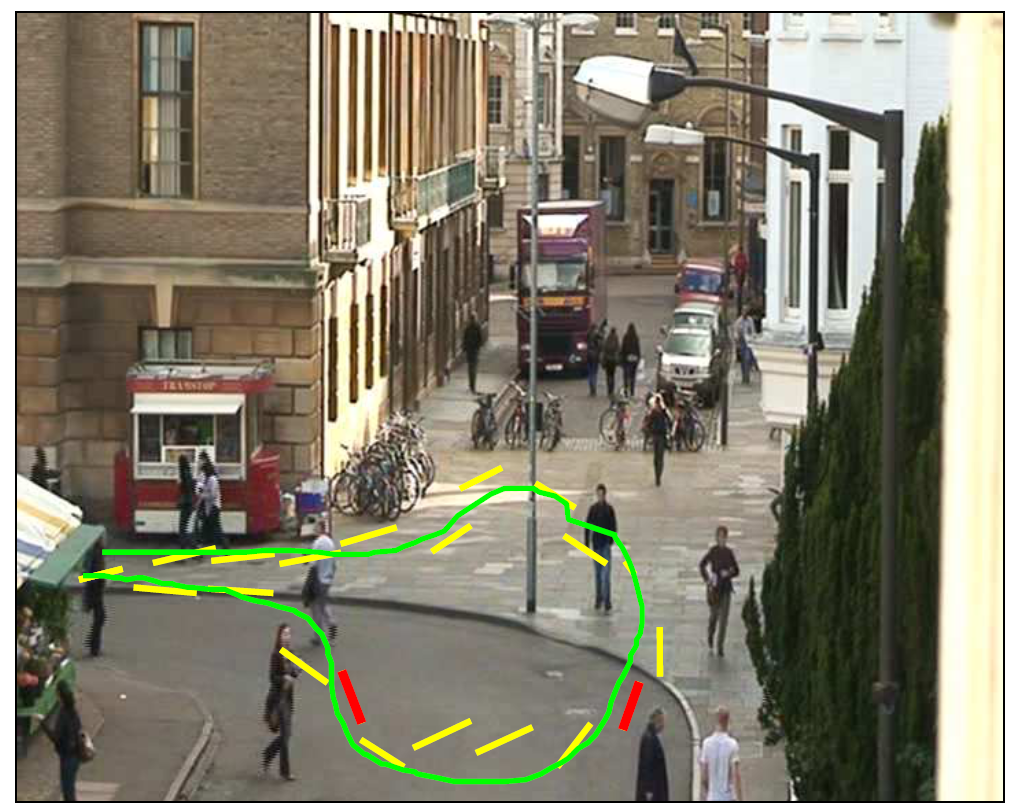}\hspace{3pt}}
  \caption{ (a) A typical frame extracted from the video footage used in the evaluation of this paper,
            showing a busy city marketplace, and (b) a magnified image region containing examples of
            multiple moving objects, mutual occlusion and occlusion by other objects in the scene
            (car and lamppost), background clutter and shadow patterns which
            change over time. (c) A synthetic track (green) which our \emph{pursuit-constrained motion model}
            based algorithm identifies as unusual, its representation as a sequence of tracklet primitives
            (yellow) and the pair of primitives (red) corresponding to the most novel behaviour
            detected in the traversed path. }
  \label{f:frame}
\end{figure}

Using a stationary camera placed on top of a small building
overlooking a busy city marketplace we recorded a video sequence
which we used to evaluate the proposed methods. This footage of the
total duration of 1h:59m:40s and spatial resolution $720 \times 576$
pixels contains all of the challenging aspects used to motivate our
work: continuous presence of a large number of moving entities,
frequent occlusions, articulation and scale changes, non-static
background and large variability in motion patterns. A typical frame
is shown in Figure~\ref{f:frame}~(a) while Figure~\ref{f:frame}~(b)
exemplifies some of the aforementioned difficulties on a magnified
subregion.

For the Markov chains ensemble model we used four different characteristic scales, with the
corresponding scale parameters $\Delta D_1 = 50,~\Delta D_2 = 75,~\Delta D_3 = 110,~\Delta D_3=150$.
The lowest scale tracklet model (with $\Delta D_1 = 50$) was used for the pursuit-constrained
motion model. The estimation of motion primitives from tracklets was performed using clustering
parameters $\Delta Q = 25$ and $\Delta \Theta = \pi/16$.

\subsection{Results}
\begin{figure}[ht]
  \centering
  \begin{tabular*}{1.00\textwidth}{@{\extracolsep{\fill}}cc}
    \multicolumn{2}{c}{Unusual track 1 \vspace{10pt}}\\
    \includegraphics[width=0.45\textwidth]{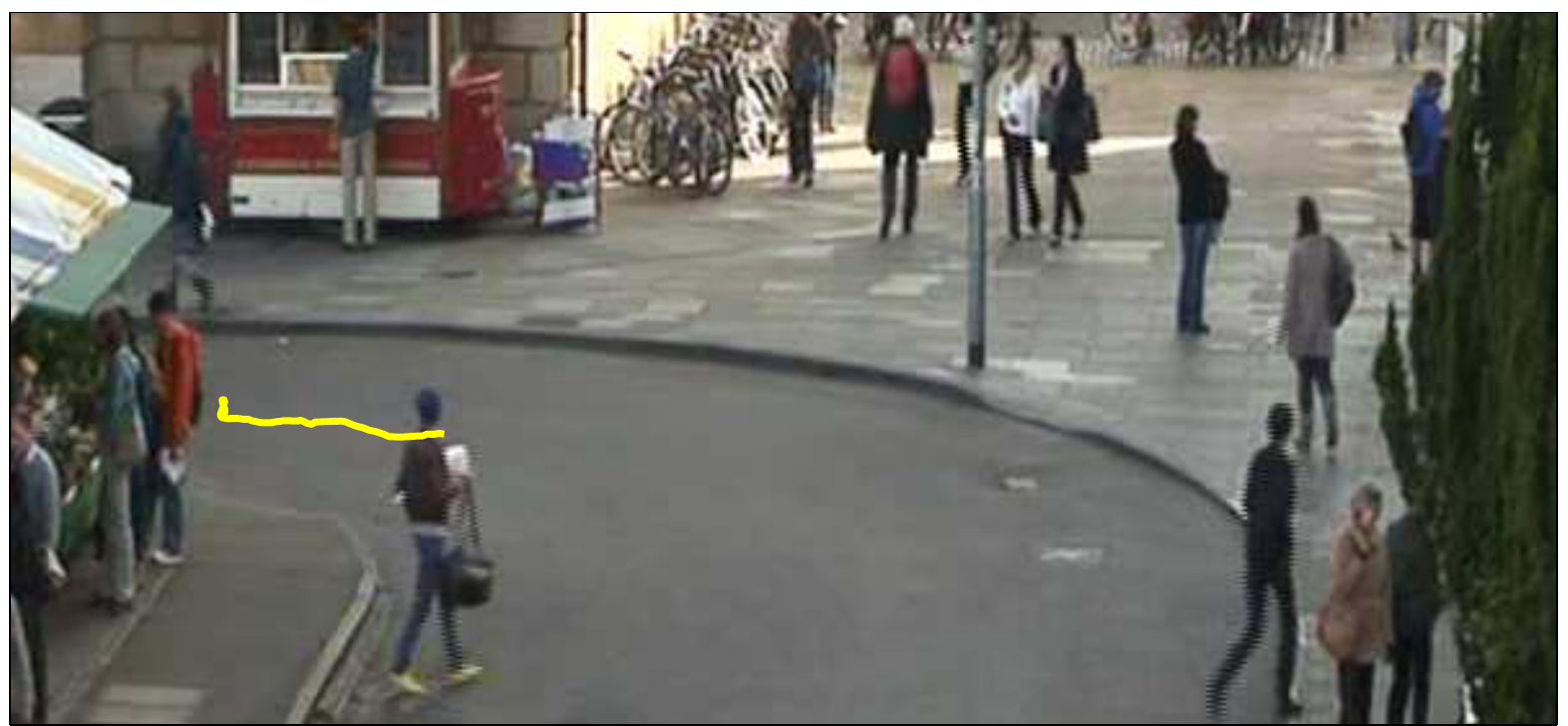}&
    \includegraphics[width=0.45\textwidth]{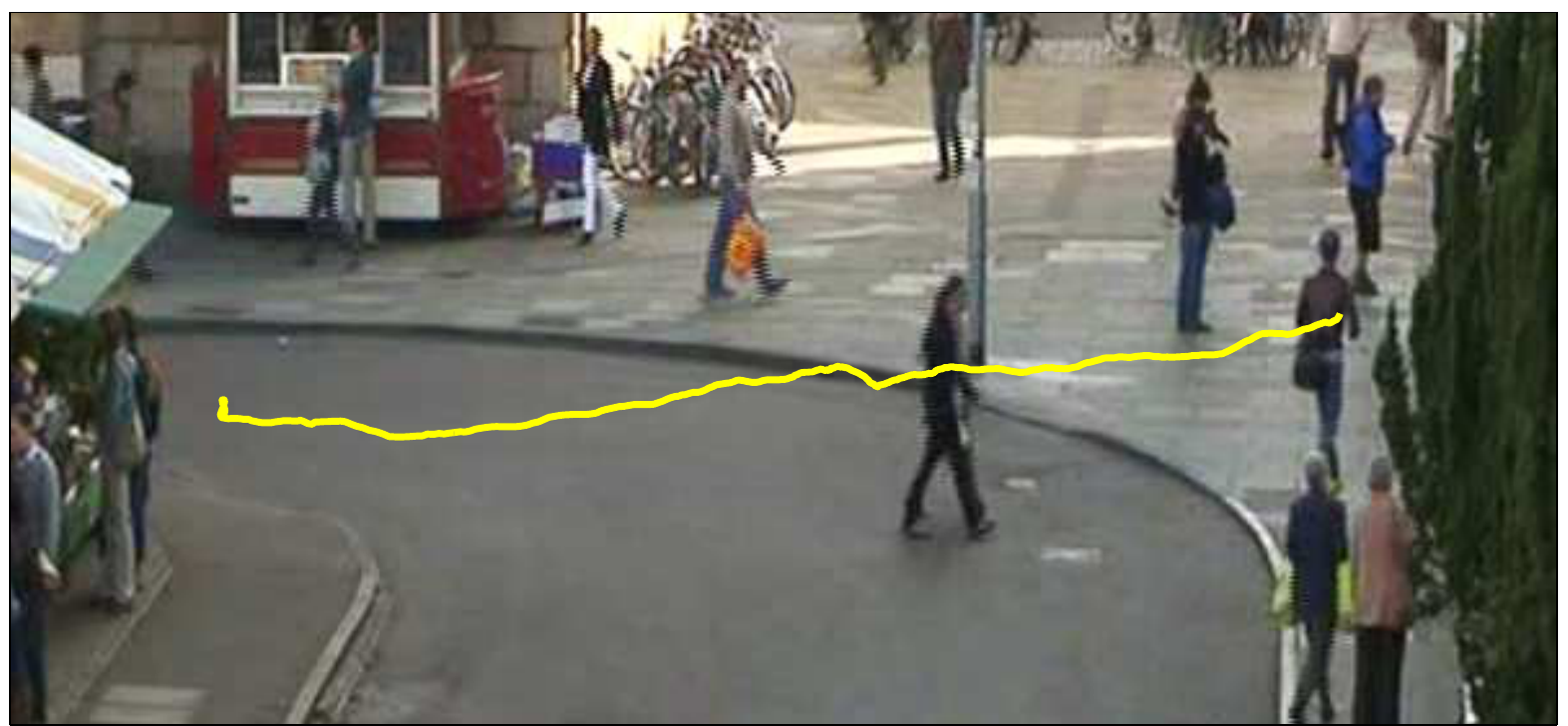}\\&\\
    \includegraphics[width=0.45\textwidth]{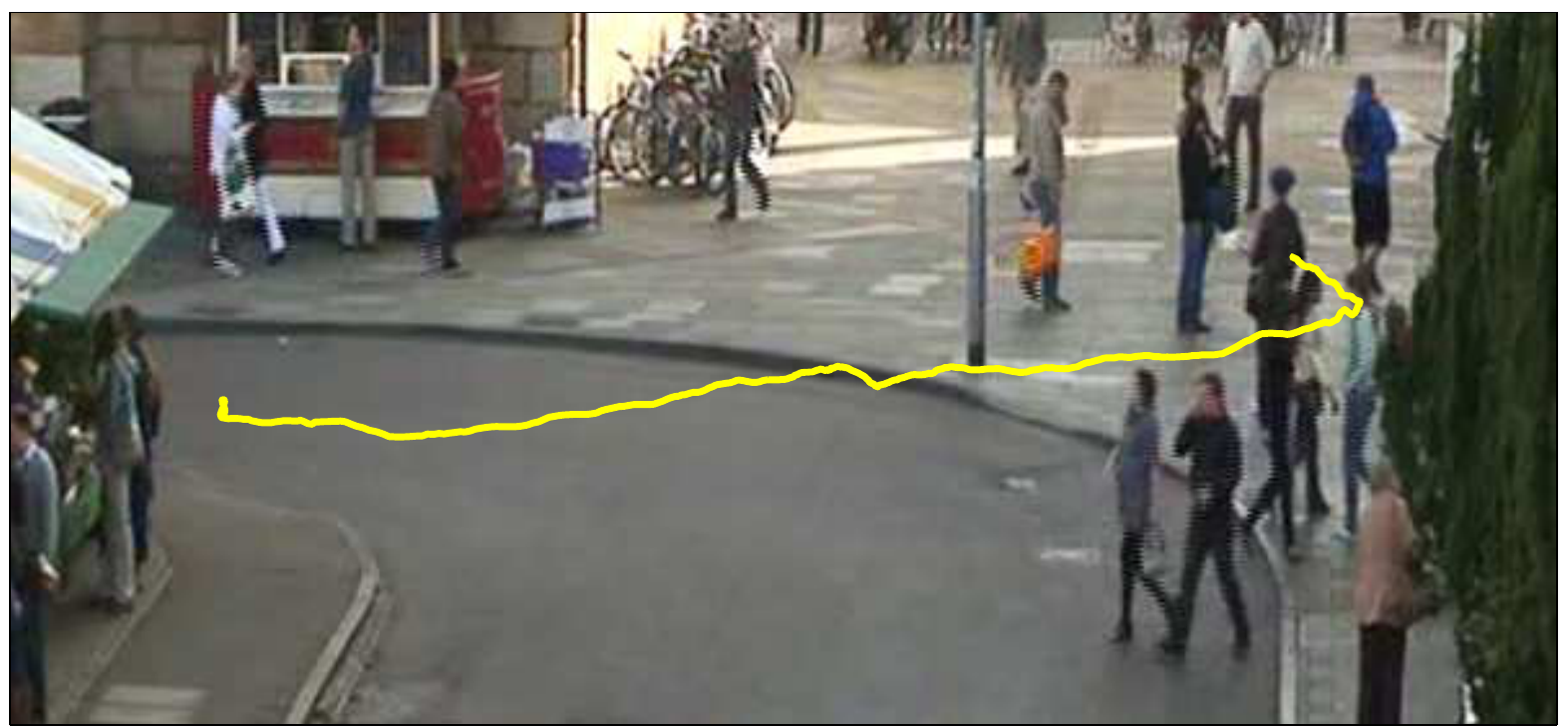}&
    \includegraphics[width=0.45\textwidth]{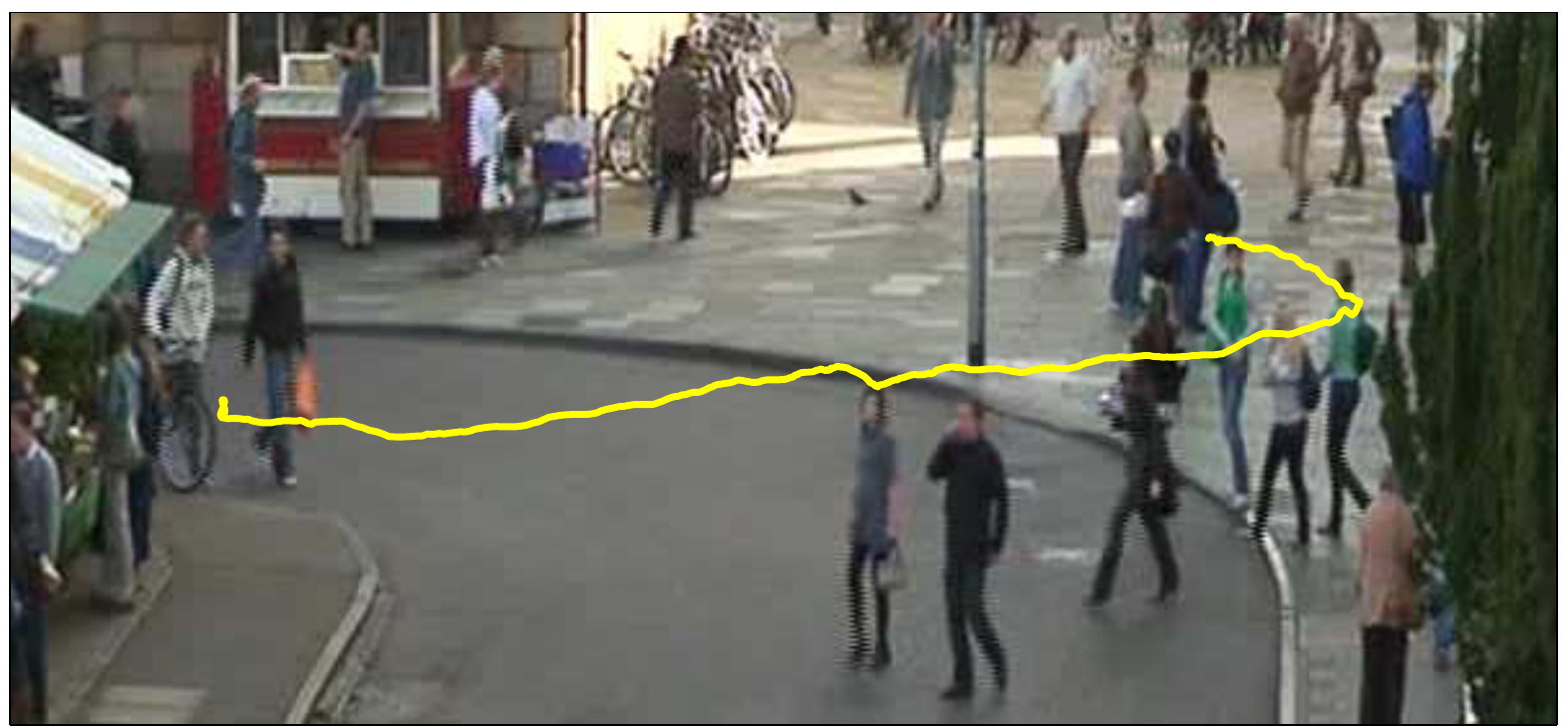}\\&\\
  \end{tabular*}
  \begin{tabular*}{1.00\textwidth}{@{\extracolsep{\fill}}ccc}
    \multicolumn{3}{c}{Unusual track 2 \vspace{10pt}}\\
    \includegraphics[width=0.31\textwidth]{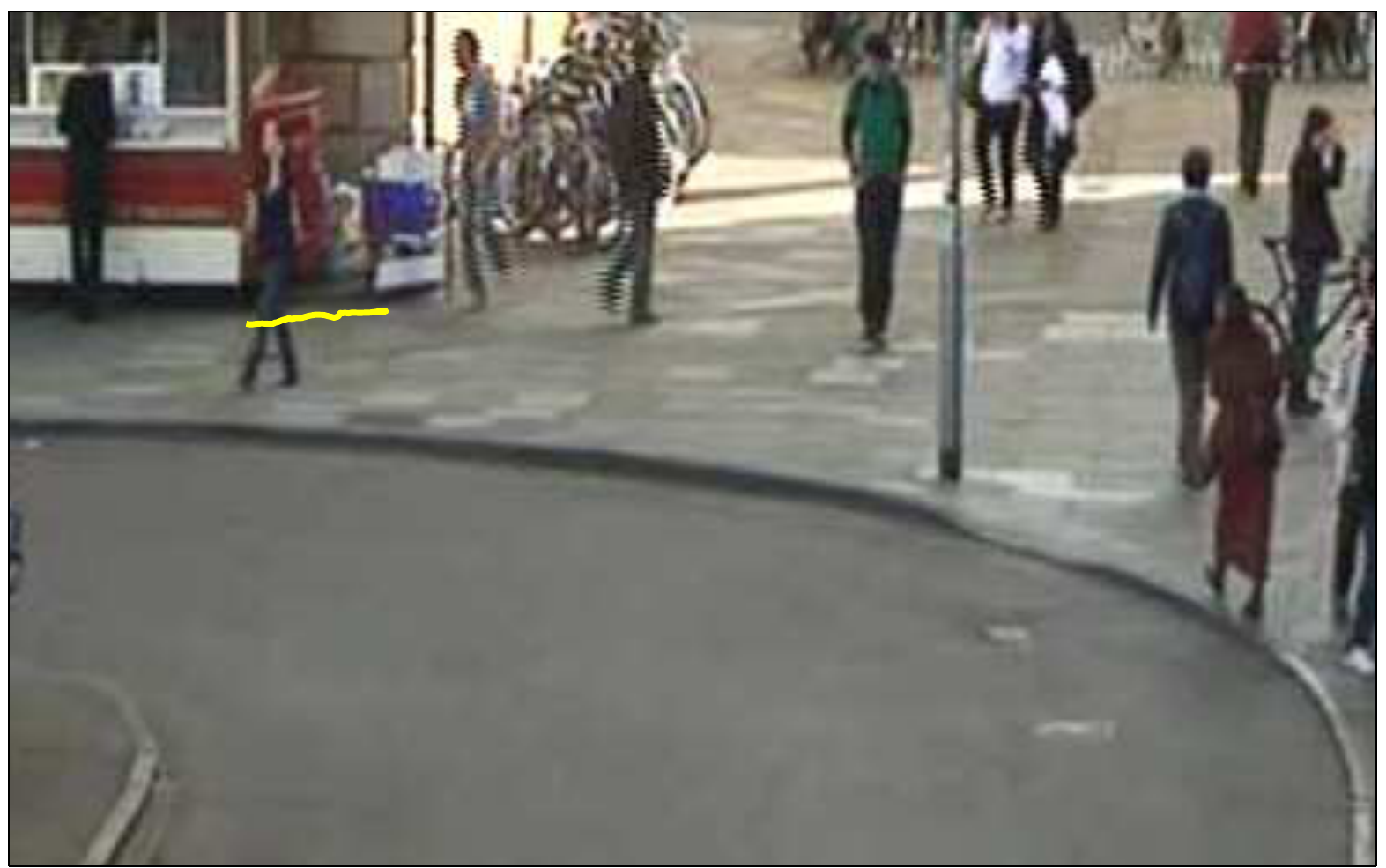}&
    \includegraphics[width=0.31\textwidth]{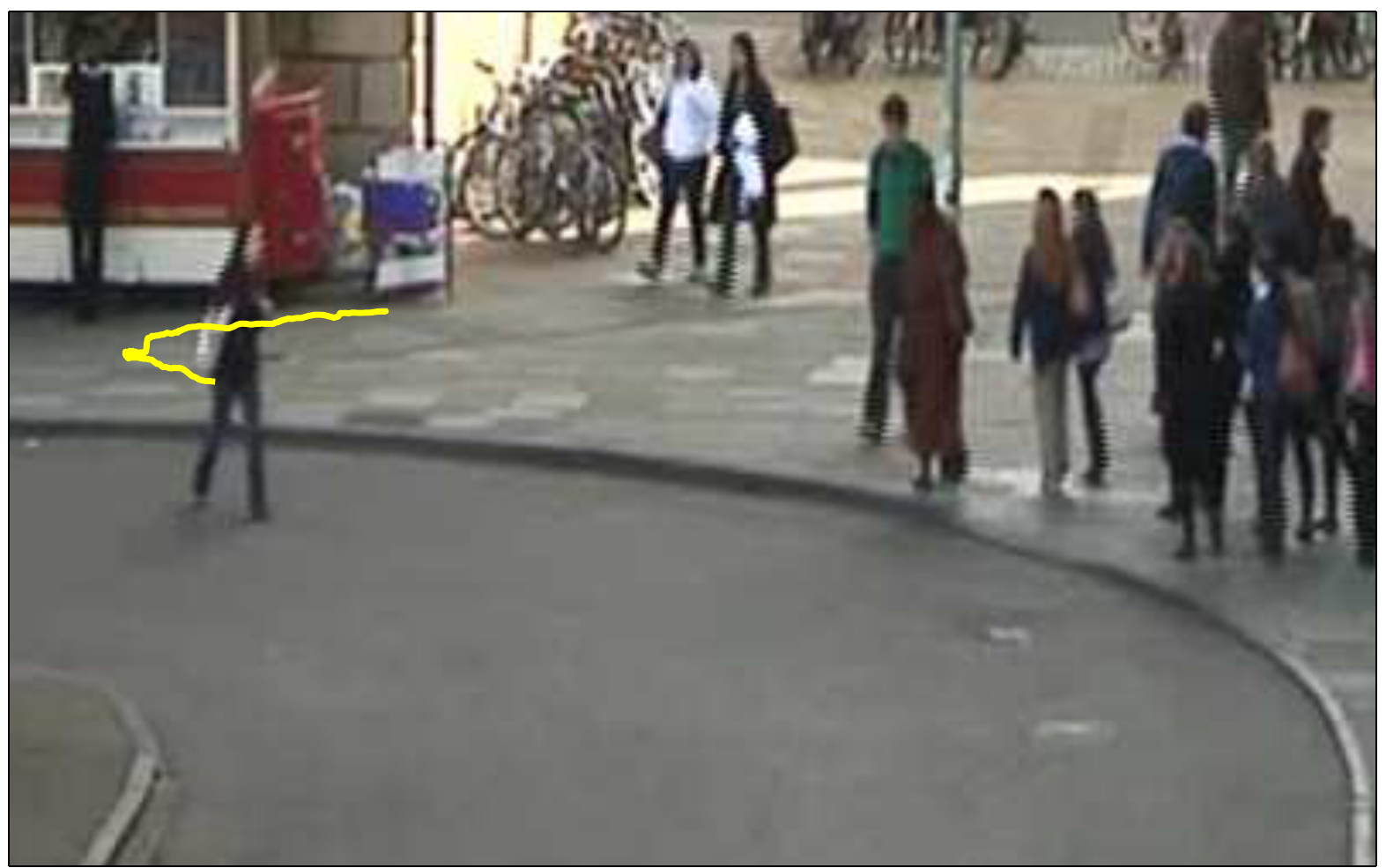}&
    \includegraphics[width=0.31\textwidth]{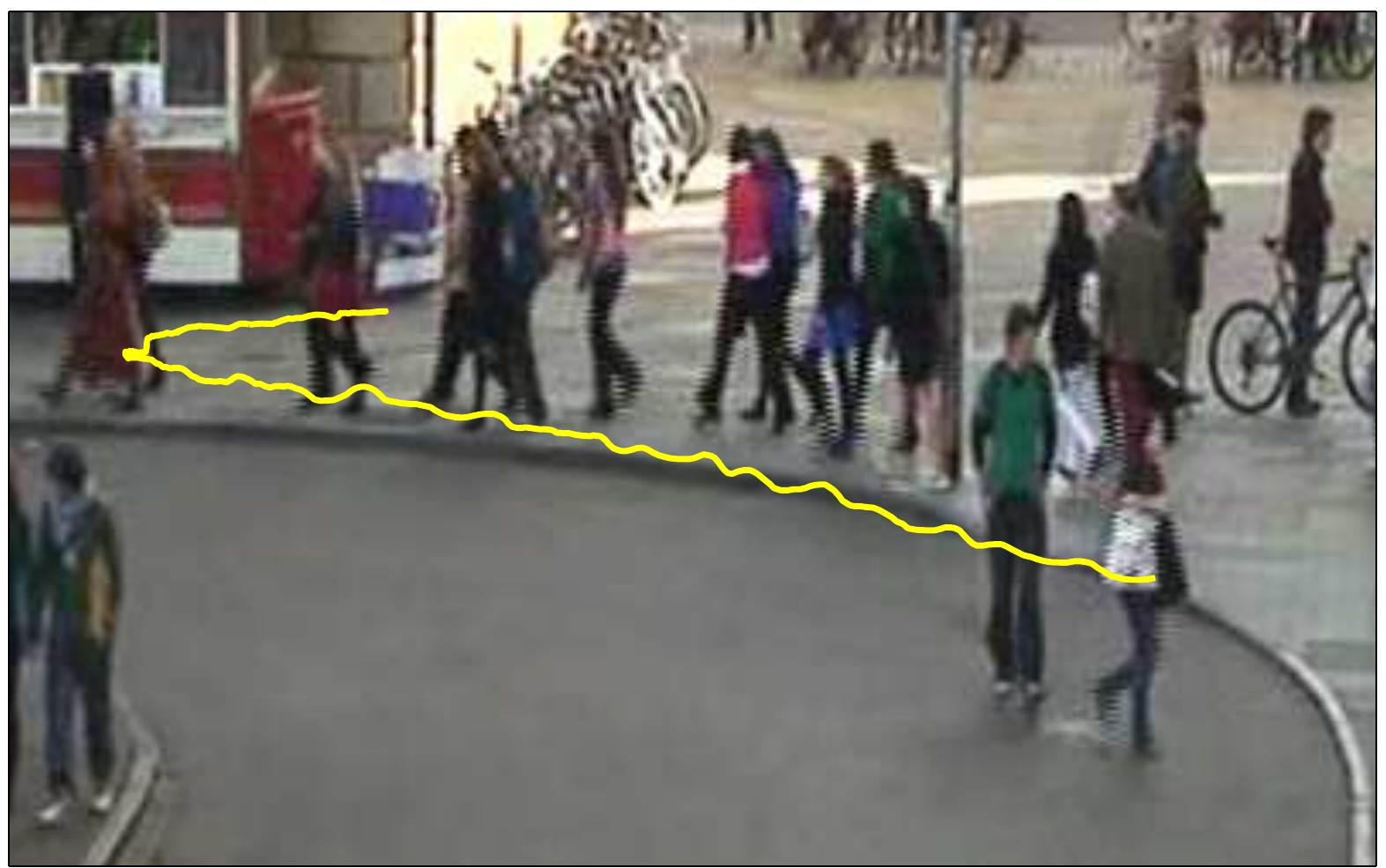}\\&&\\
    \multicolumn{3}{c}{Unusual track 3 \vspace{10pt}}\\
    \includegraphics[width=0.25\textwidth]{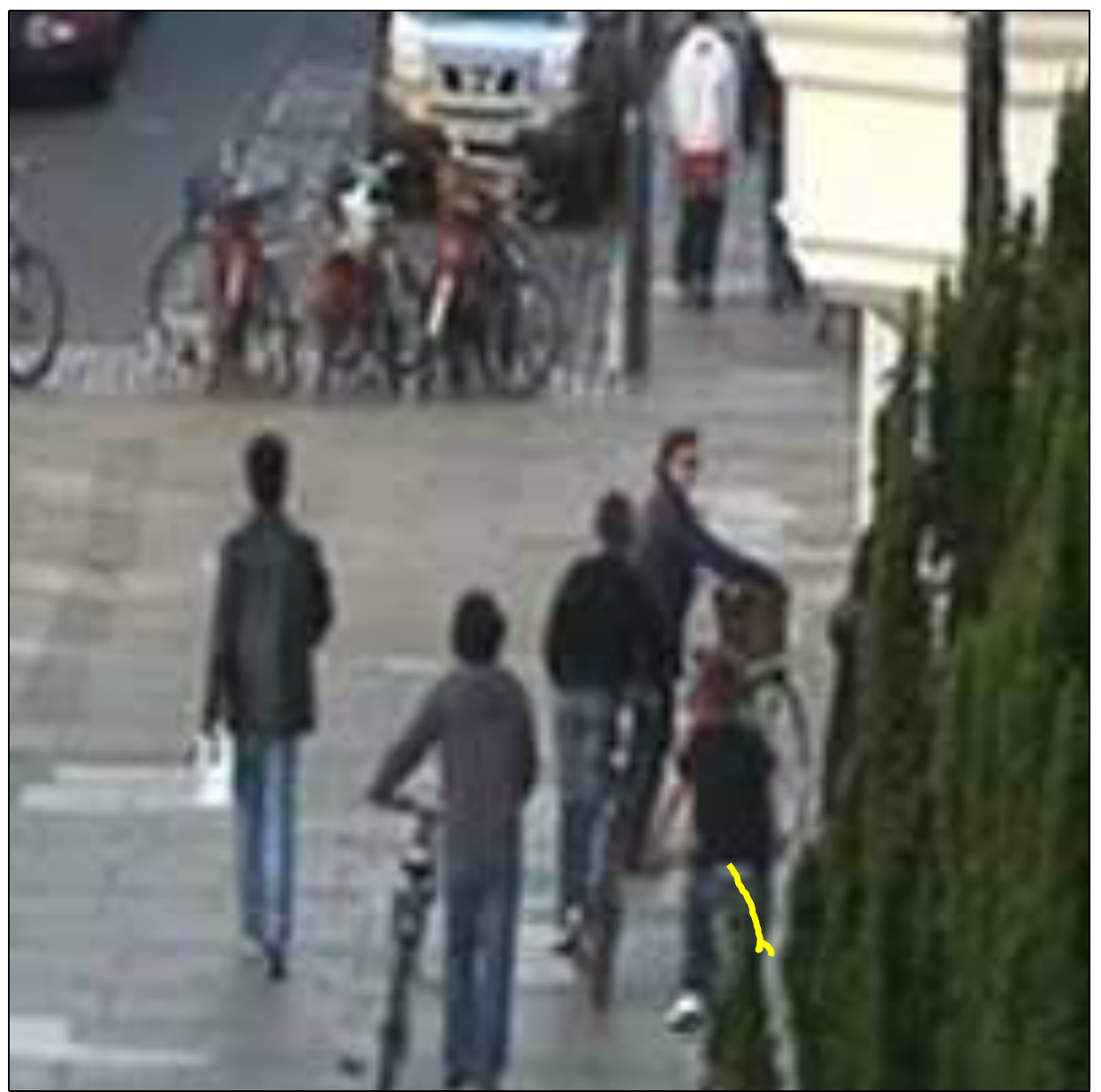}&
    \includegraphics[width=0.25\textwidth]{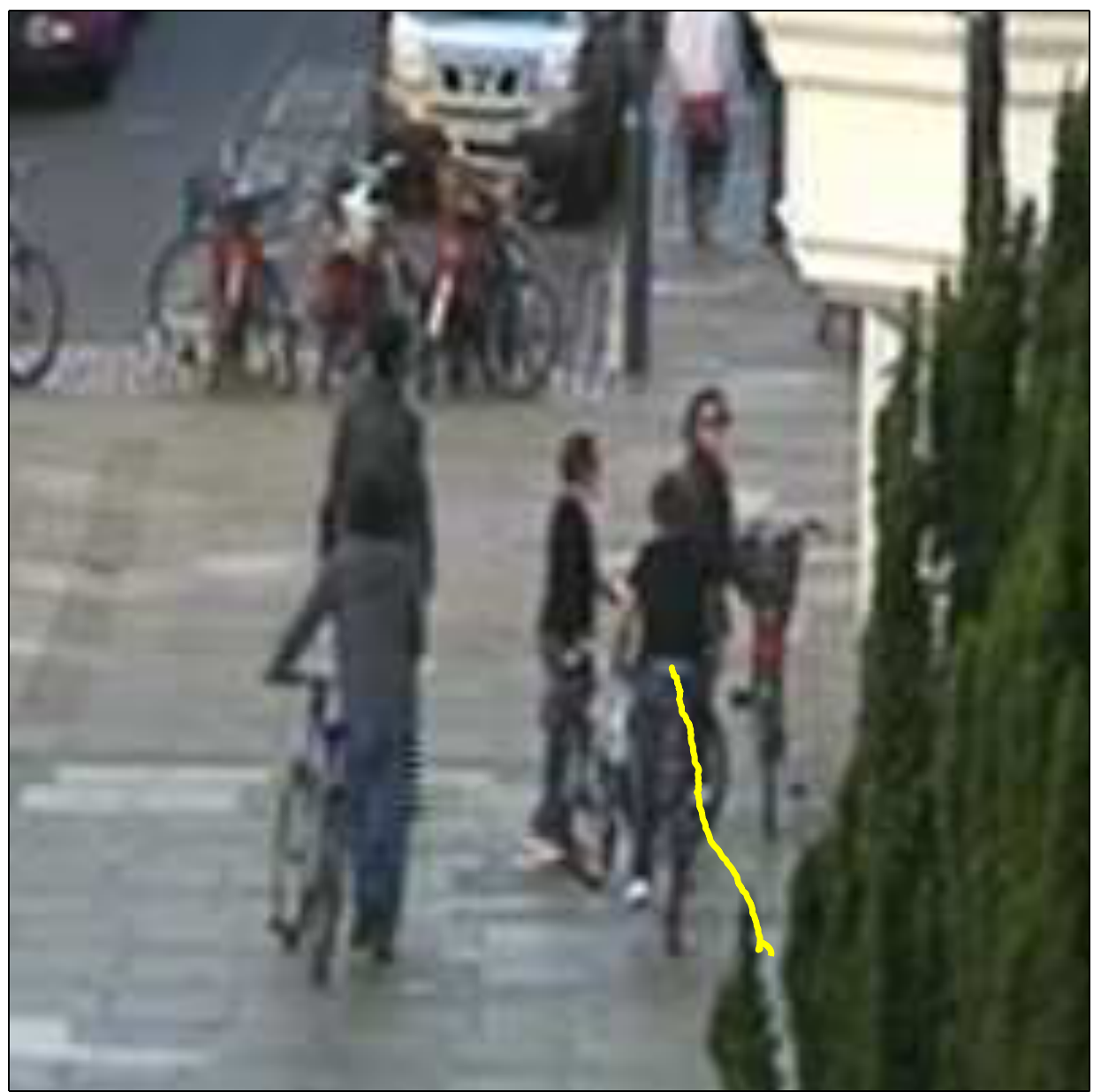}&
    \includegraphics[width=0.25\textwidth]{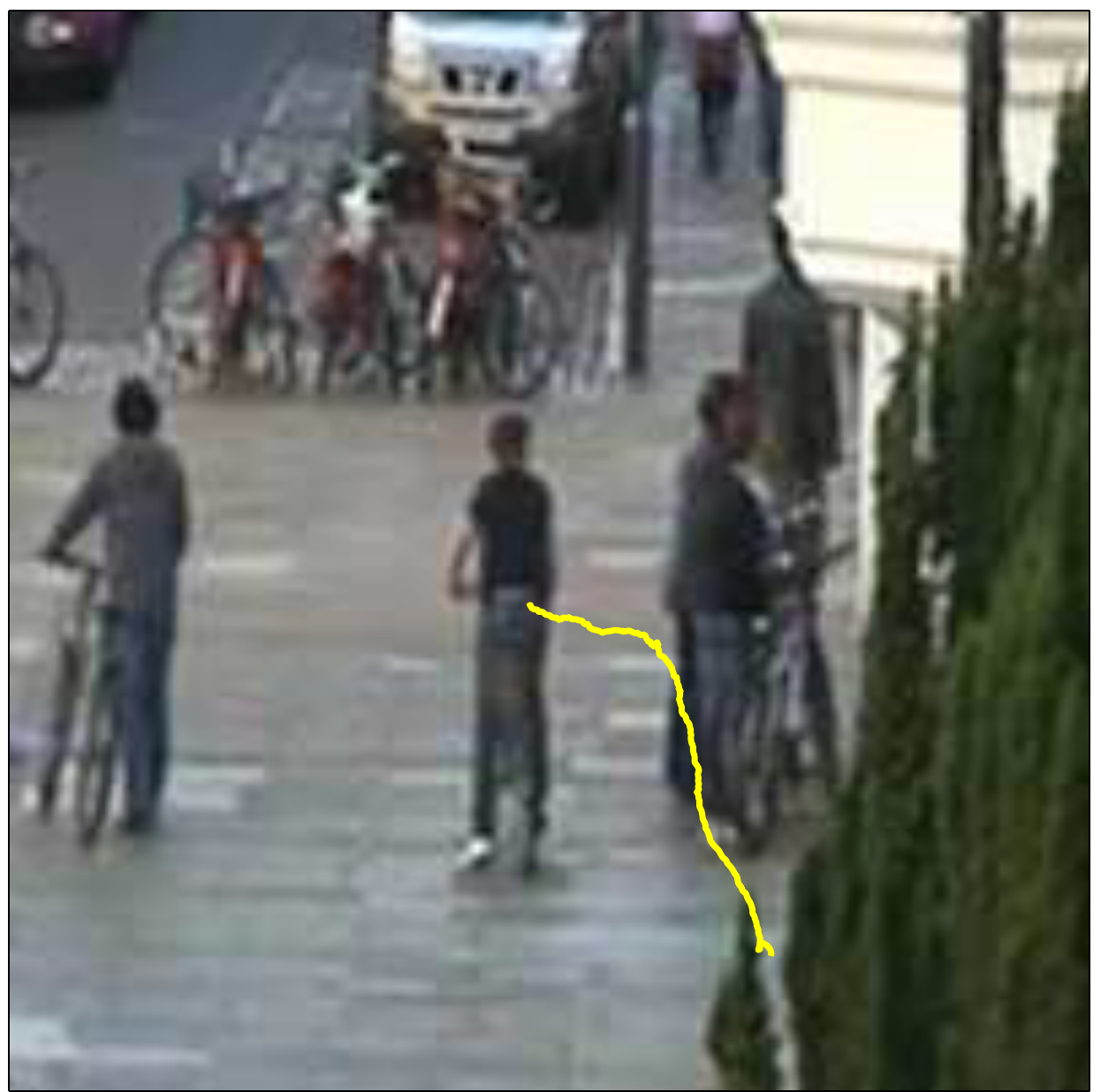}\\&&\\
    \includegraphics[width=0.25\textwidth]{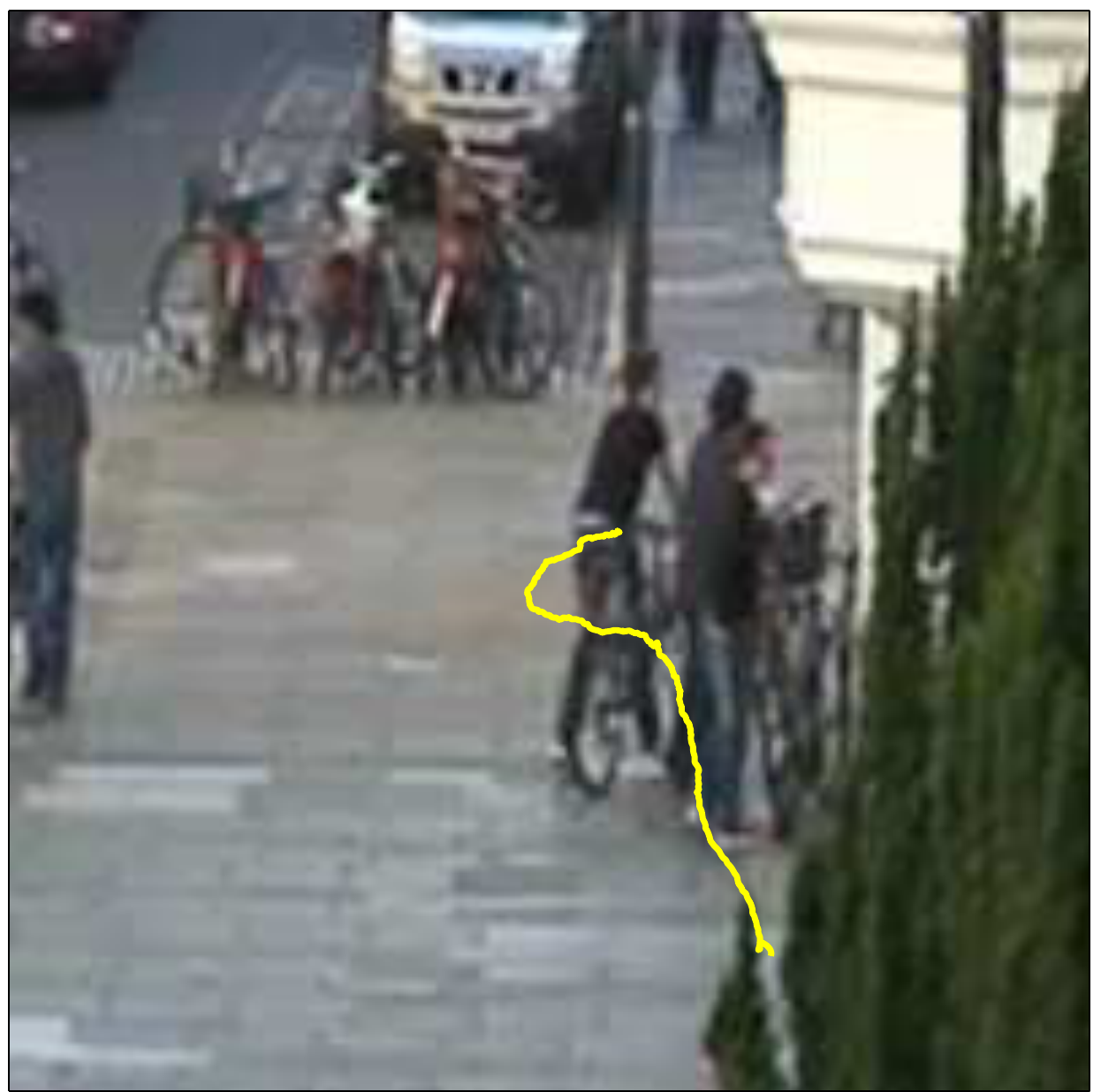}&
    \includegraphics[width=0.25\textwidth]{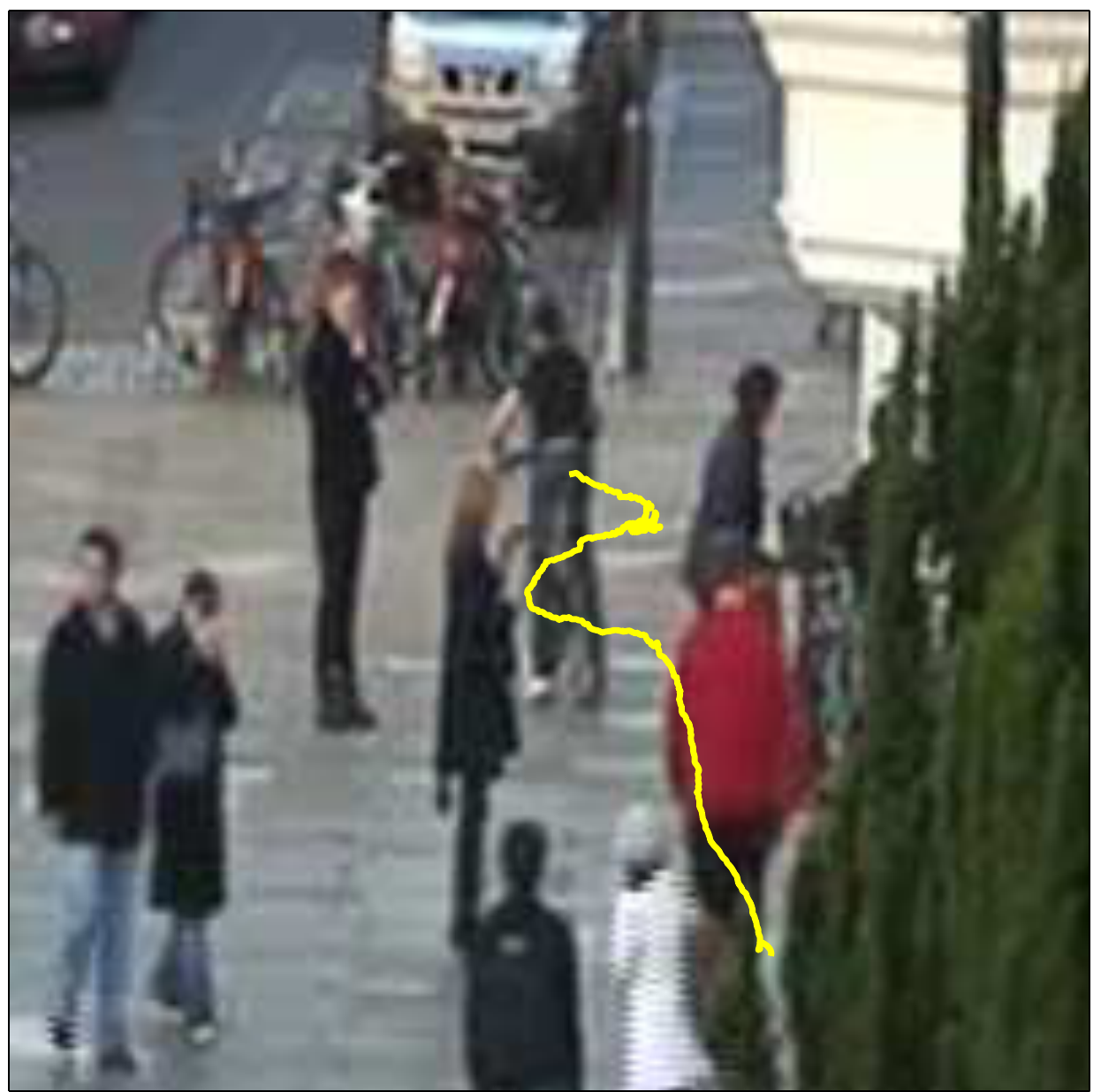}&
    \includegraphics[width=0.25\textwidth]{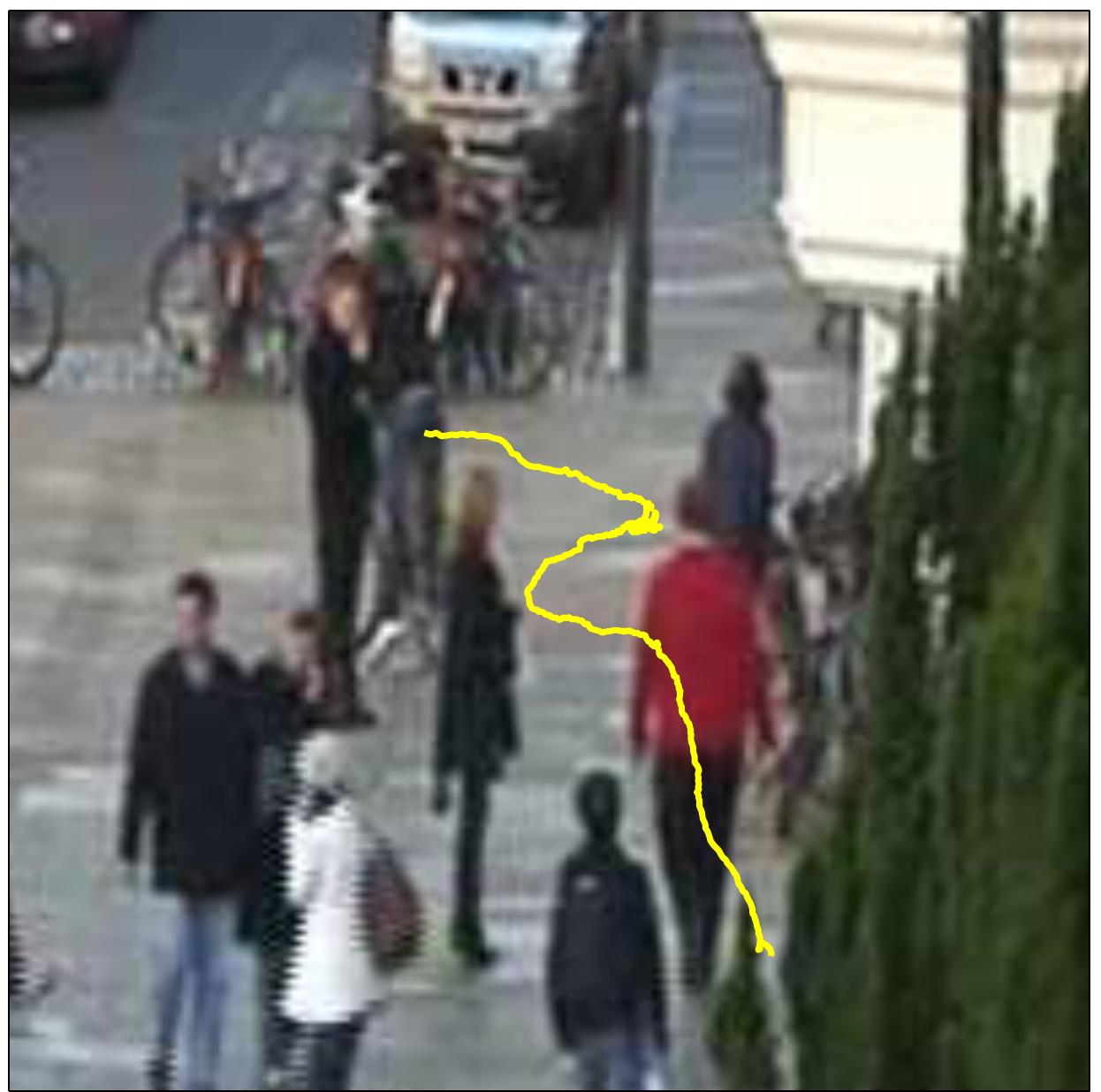}\\
  \end{tabular*}
  \caption{ Examples of behaviour detected as unusual by our algorithms. Also see Figure~\ref{f:res2}. }
  \label{f:res1}
\end{figure}

\begin{figure}[ht]
  \centering
  \begin{tabular*}{1.00\textwidth}{@{\extracolsep{\fill}}cc}
    \multicolumn{2}{c}{Unusual track 4 \vspace{10pt}}\\
    \includegraphics[width=0.45\textwidth]{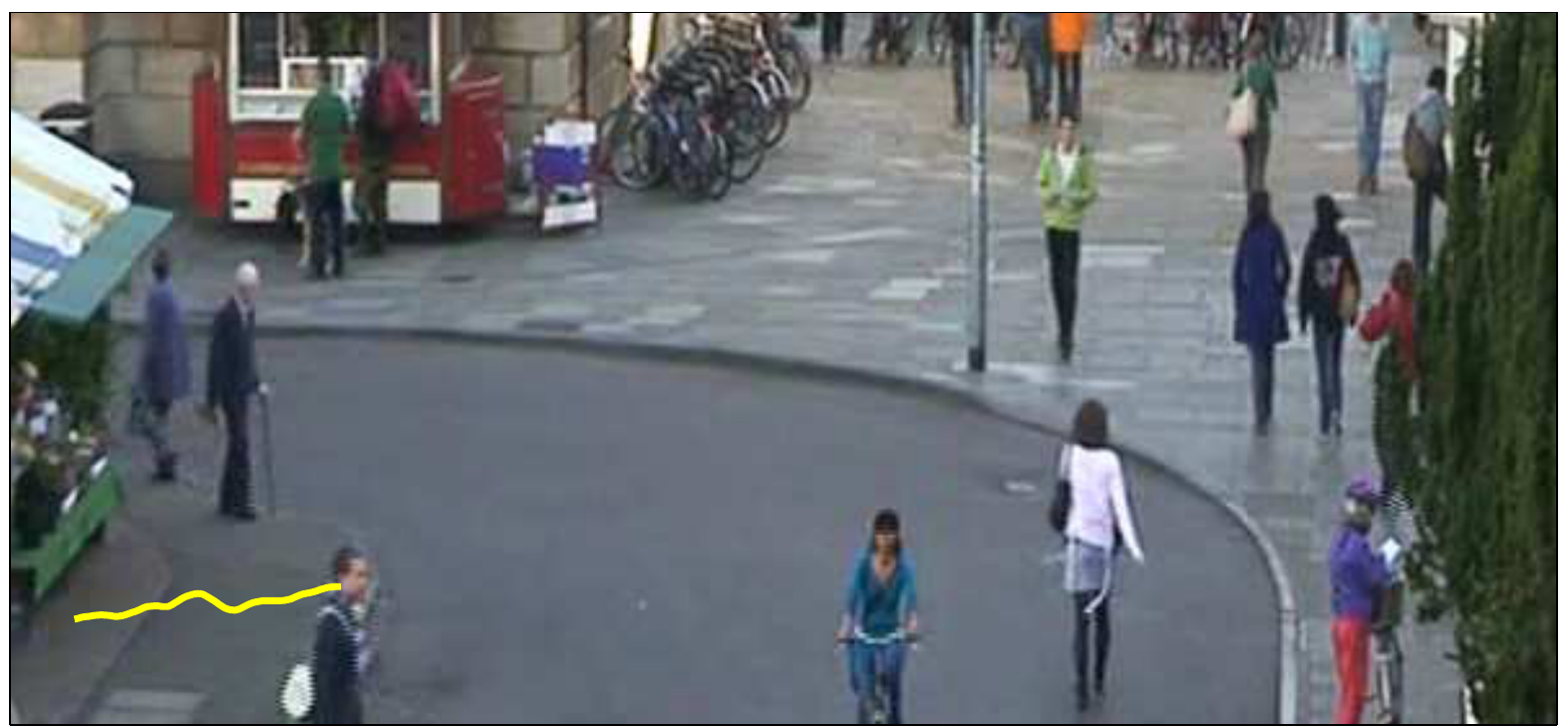}&
    \includegraphics[width=0.45\textwidth]{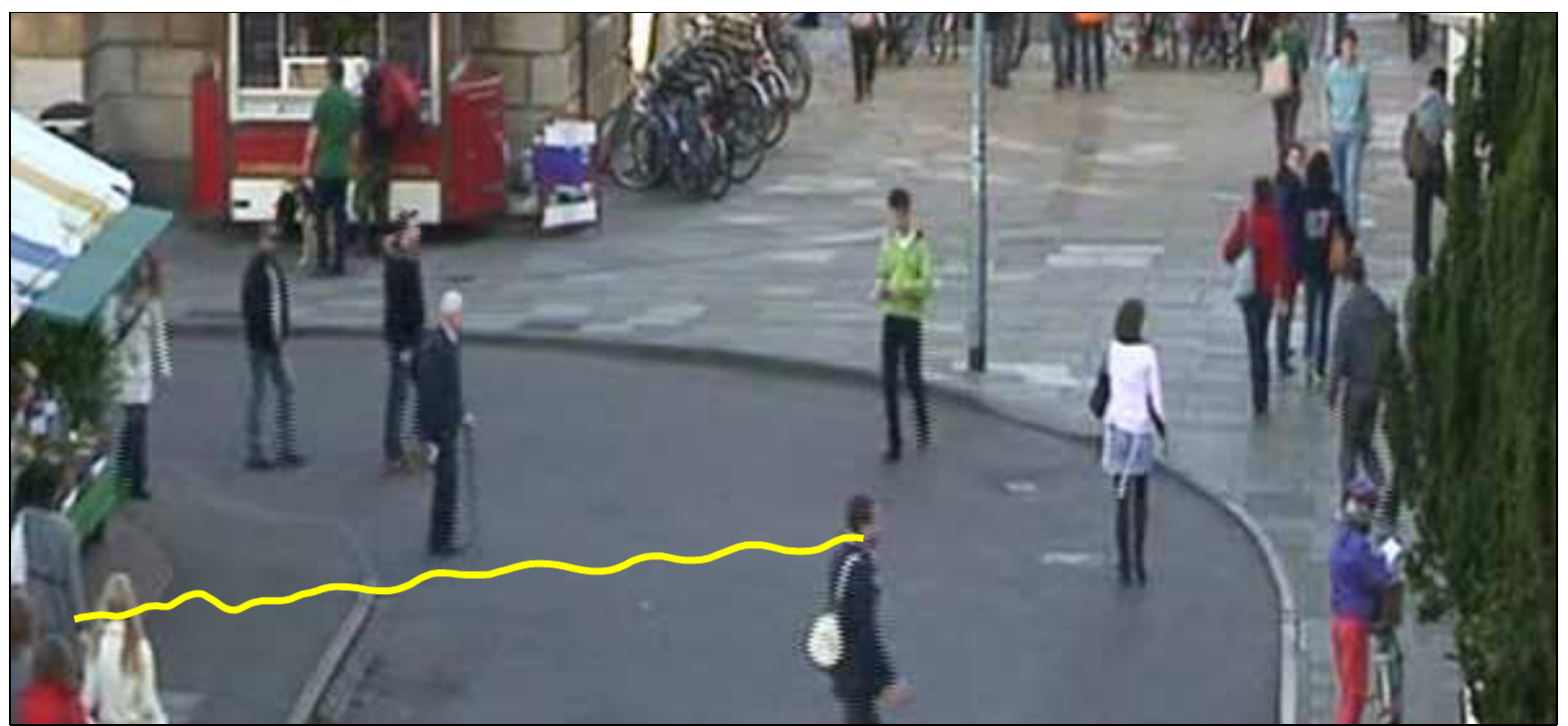}\\&\\
    \includegraphics[width=0.45\textwidth]{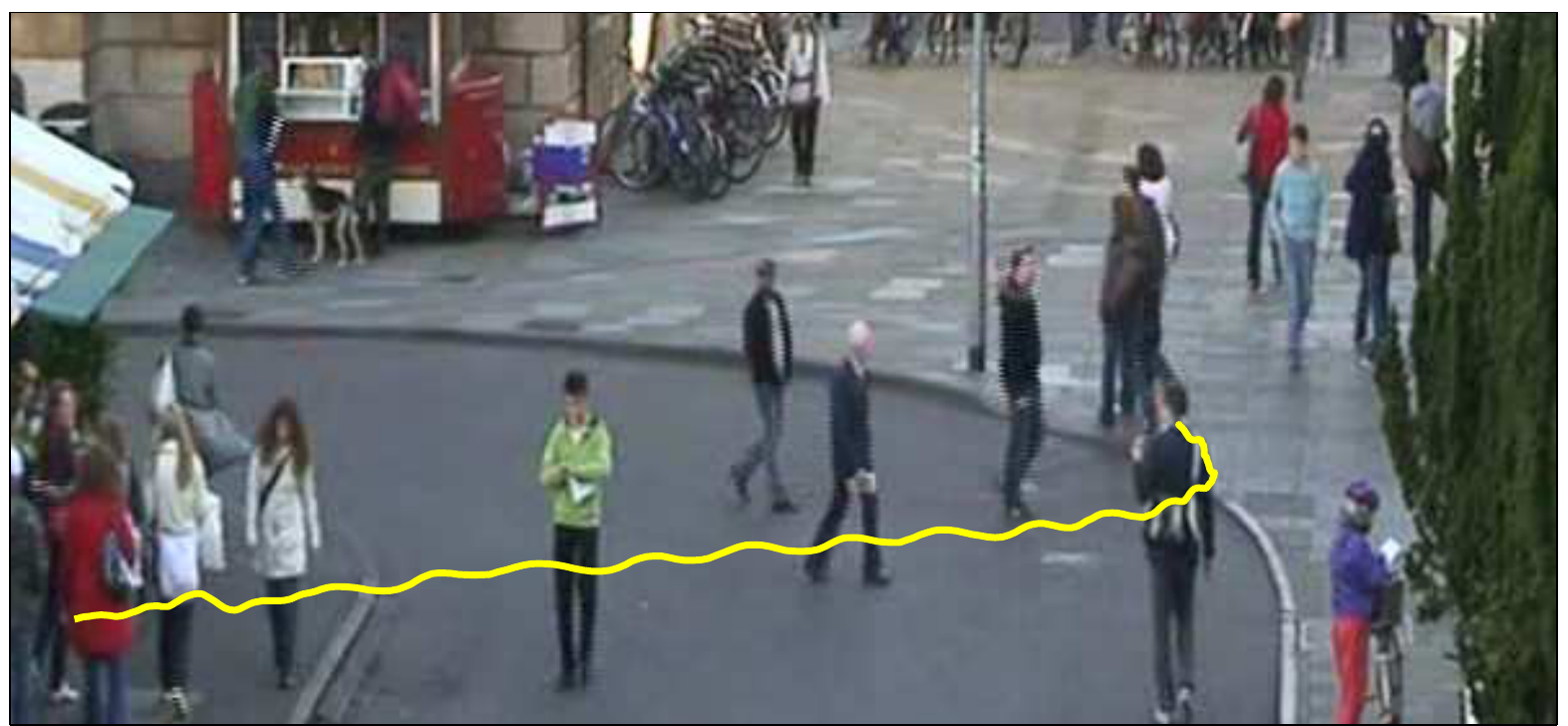}&
    \includegraphics[width=0.45\textwidth]{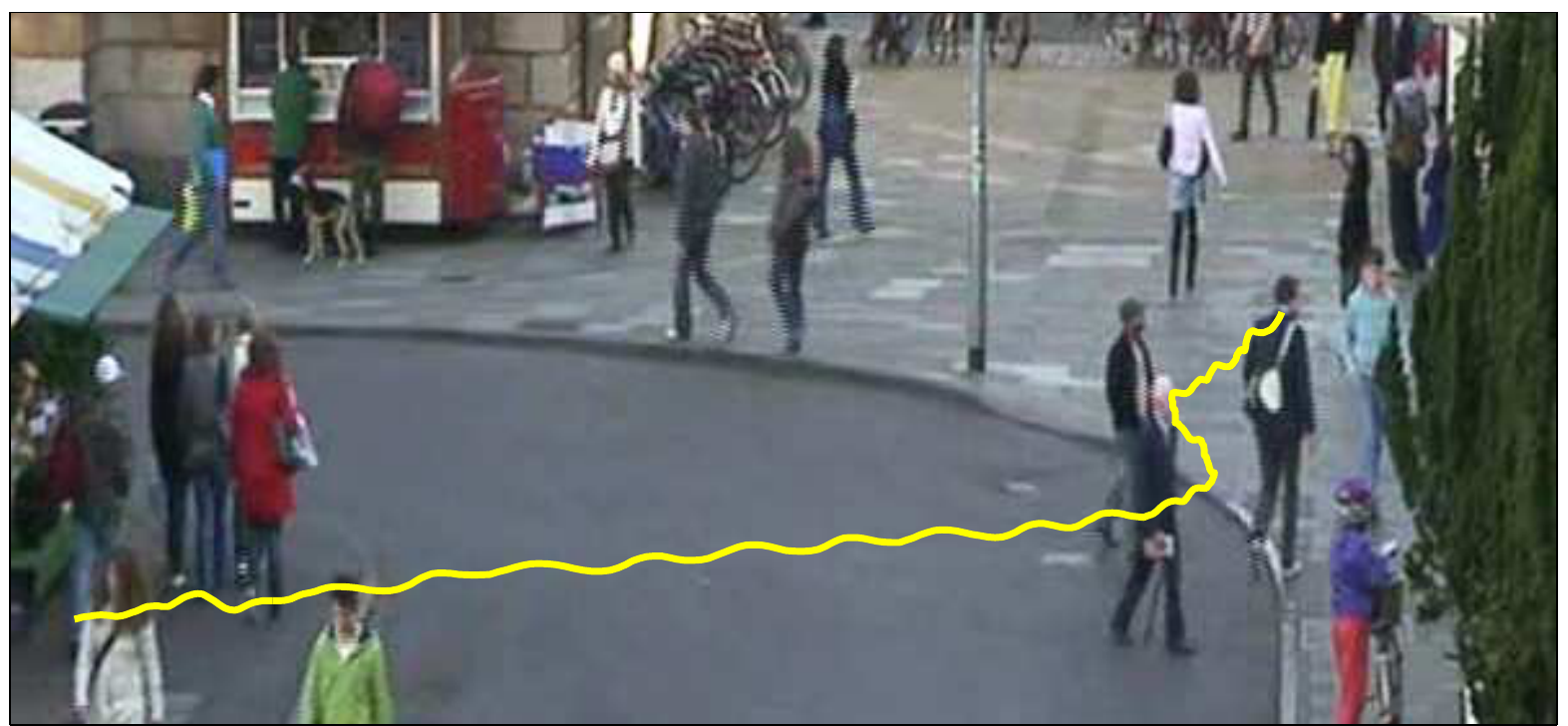}\\&\\
    \includegraphics[width=0.45\textwidth]{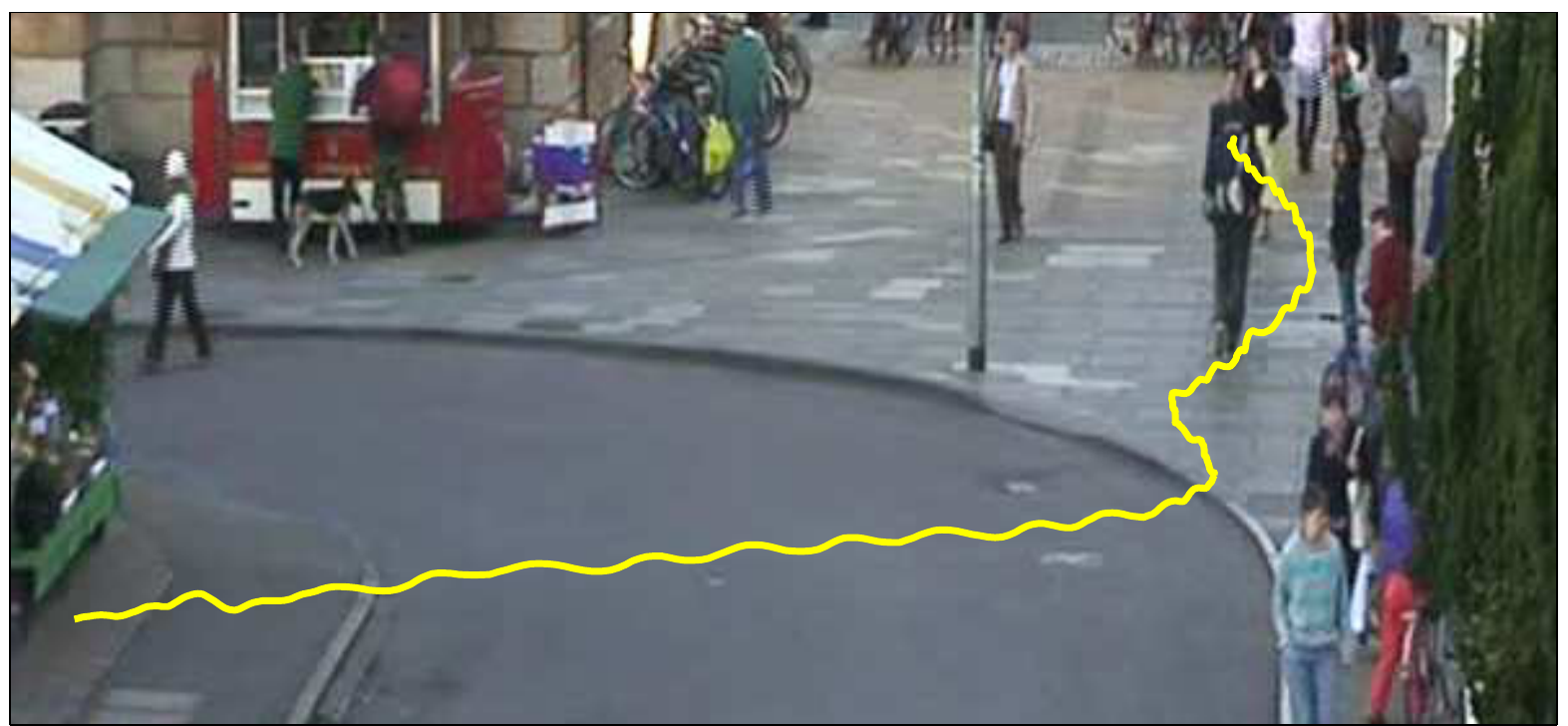}&
    \includegraphics[width=0.45\textwidth]{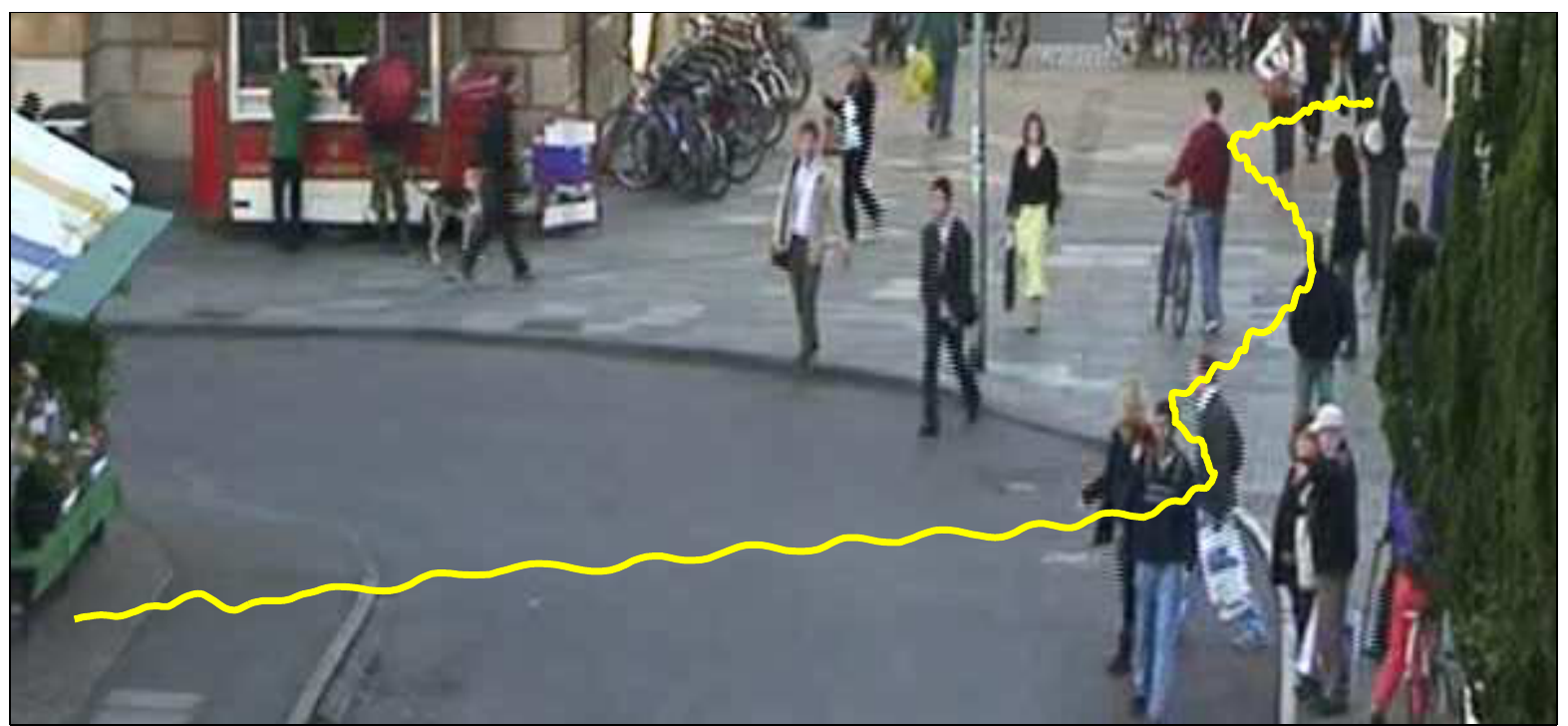}\\&\\
  \end{tabular*}
  \begin{tabular*}{1.00\textwidth}{@{\extracolsep{\fill}}cc}
    \multicolumn{2}{c}{Unusual track 5 \vspace{10pt}}\\
    \includegraphics[width=0.45\textwidth]{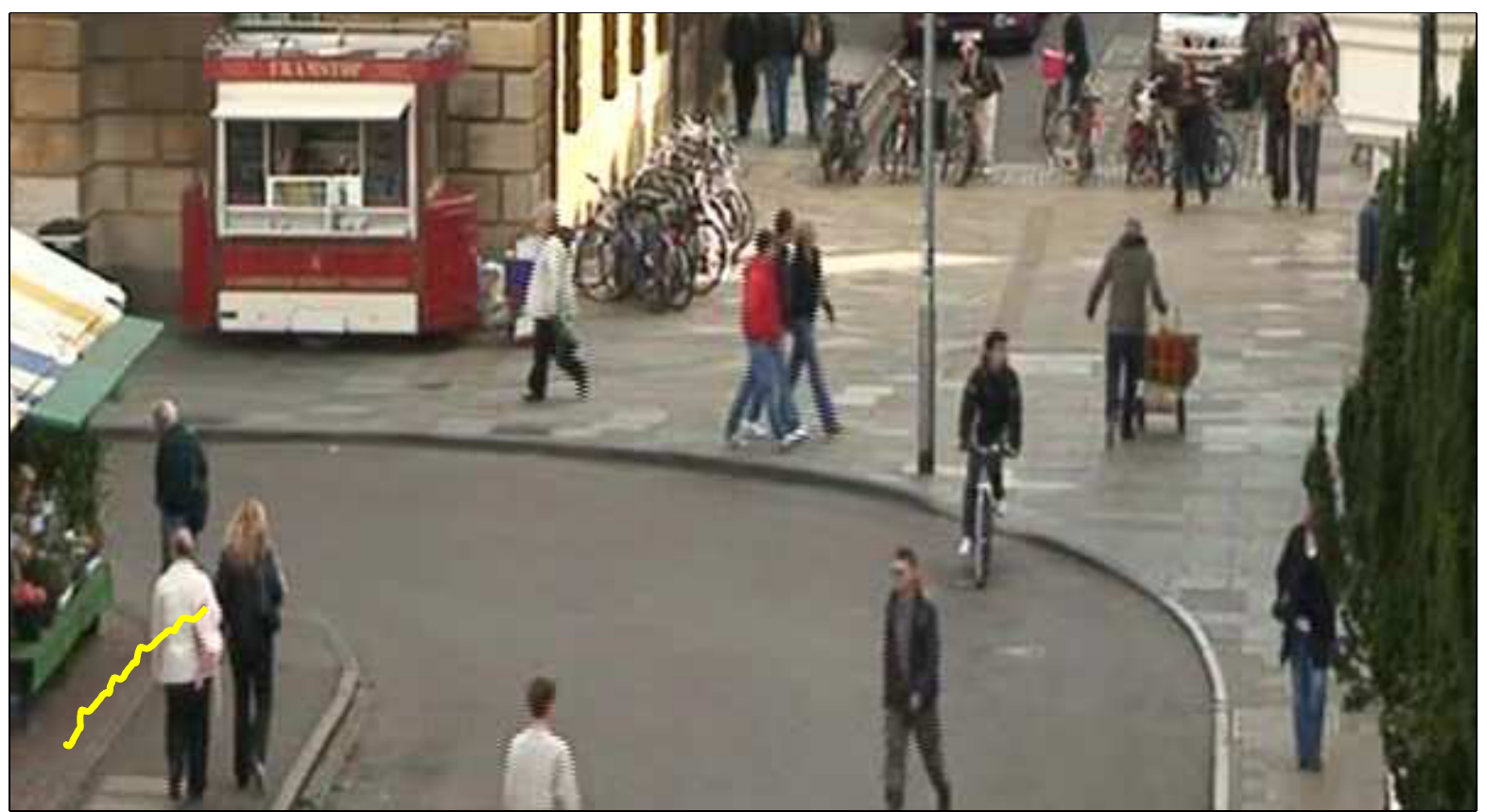}&
    \includegraphics[width=0.45\textwidth]{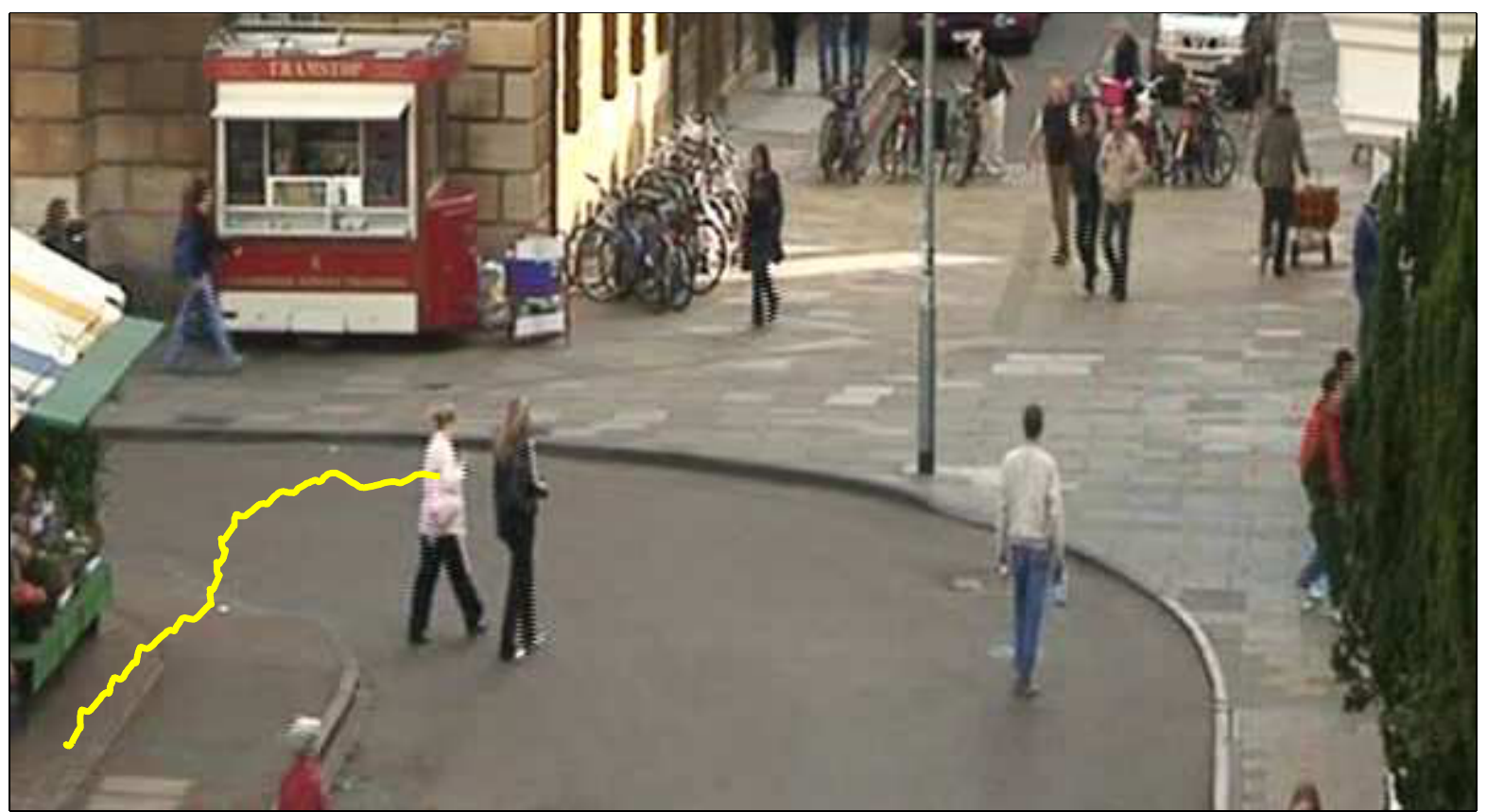}\\&\\
    \includegraphics[width=0.45\textwidth]{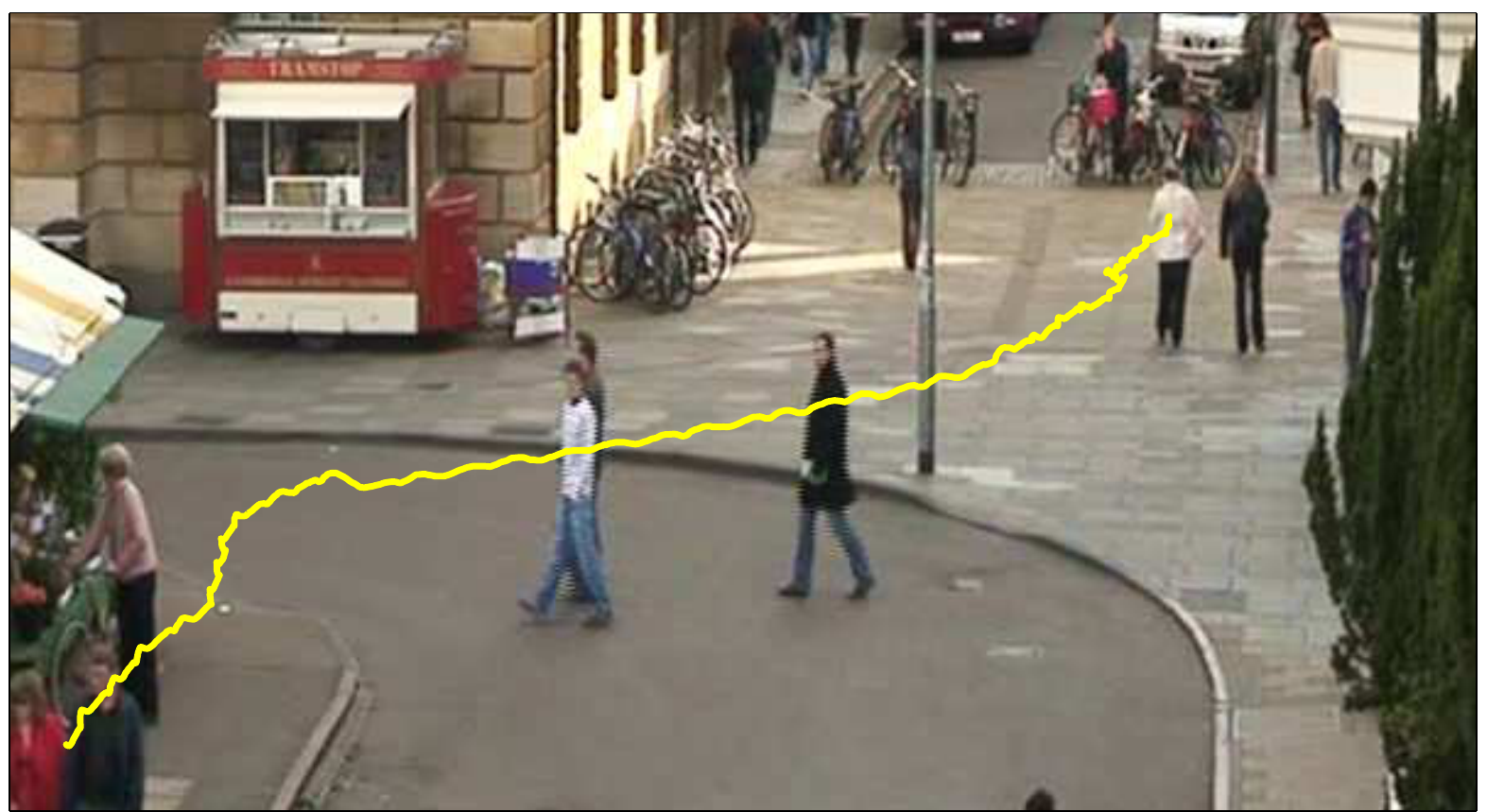}&\\
  \end{tabular*}
  \caption{ Examples of behaviour detected as unusual by our algorithms. Also see Figure~\ref{f:res1}. }
  \label{f:res2}
\end{figure}

After training each of the proposed methods using the 113,700
extracted tracks, we computed the corresponding histograms of
conformity measures $\rho_1$ and $\rho_2$. From these we
automatically selected thresholds for novelty detection, $R_1$ and
$R_2$, such that 0.05\% of training tracks produce lower
conformities. An examination of these tracks revealed that the two
algorithms generally identified the same tracks as being the least
like the rest of the training set with several typical results shown
in Figures~\ref{f:res1} and~\ref{f:res2}. A common aspect which can
be observed between them is that they correspond to motions which
include sharp direction changes \emph{at locations in the scene
where there is little reason for them}. This motion not only novel
by the construction of our model but it also conforms to our
intuition about what constitutes unusual behaviour. Note the
scene-specific, contextual aspect of the learnt motions: many
extracted tracks contain sharp turns (e.g.\ at the end of the row of
marketplace stalls or at the corner of the buildings) which are not
deemed unusual because the constraints of the scene made such turns
(comparatively) frequent in the training data.

\begin{figure}[t]
  \centering
  \includegraphics[width=1.00\textwidth]{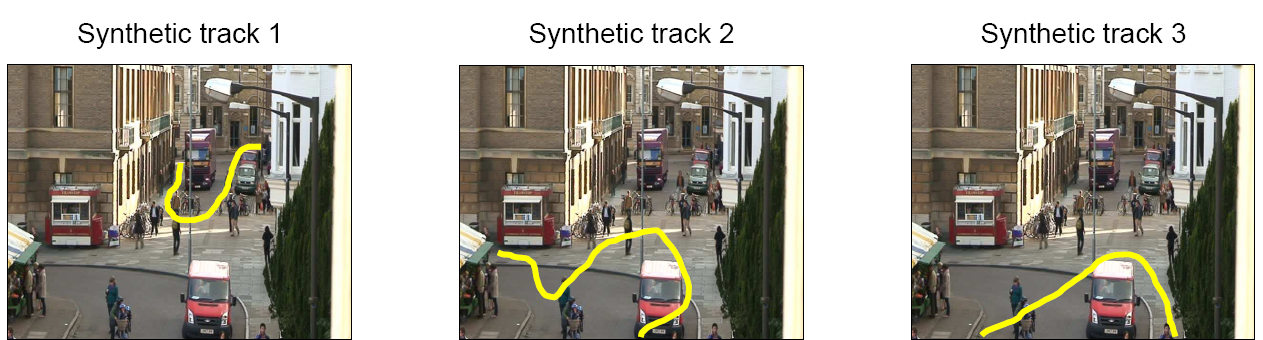}
  \caption{ Examples of three synthetic tracks which the Markov chains ensemble model
            does not classify as novel, and the \emph{pursuit-constrained motion model} does.  }
  \label{f:synth}
\end{figure}

Next, to evaluate how our algorithms cope with unseen data, by clicking on the
image of the scene we generated a series of synthetic tracks which a human might consider unusual
in the context of the marketplace in question. Here, the two methods produced different results.
Specifically, consider the examples shown in Figure~\ref{f:synth}, which the Markov chains ensemble
does not classify as novel, unlike the pursuit-constrained motion model. The discrepancy can be
explained by observing that in the ensemble approach, a trade-off is made between the precision of
motion localization by tracklet primitives and the ability to capture behaviour at a larger scale.
Such compromise does not exist in the proposed pursuit-constrained model.

Lastly, since the conformance measure $\rho_2$ in \eqref{e:rho2} is effectively dependent
only on a single pair of primitives (those which are explained the worst by the path length model),
we visualized these for a series of tracks in which novelty was detected. This is useful as a way
of ensuring that the model is capturing
meaningful information and potentially in practice as well, by drawing attention not only to a
particular behaviour on the whole but a particular feature. This is illustrated with an example
in Figure~\ref{f:frame}~(c) which shows a
synthetic track identified as unusual by the pursuit-constrained model, its decomposition into
tracklet primitives and the pair of primitives corresponding to the track's conformance score.

\begin{figure}[t]
  \centering
  \begin{tabular*}{1.00\textwidth}{@{\extracolsep{\fill}}cc}
    \includegraphics[width=0.45\textwidth]{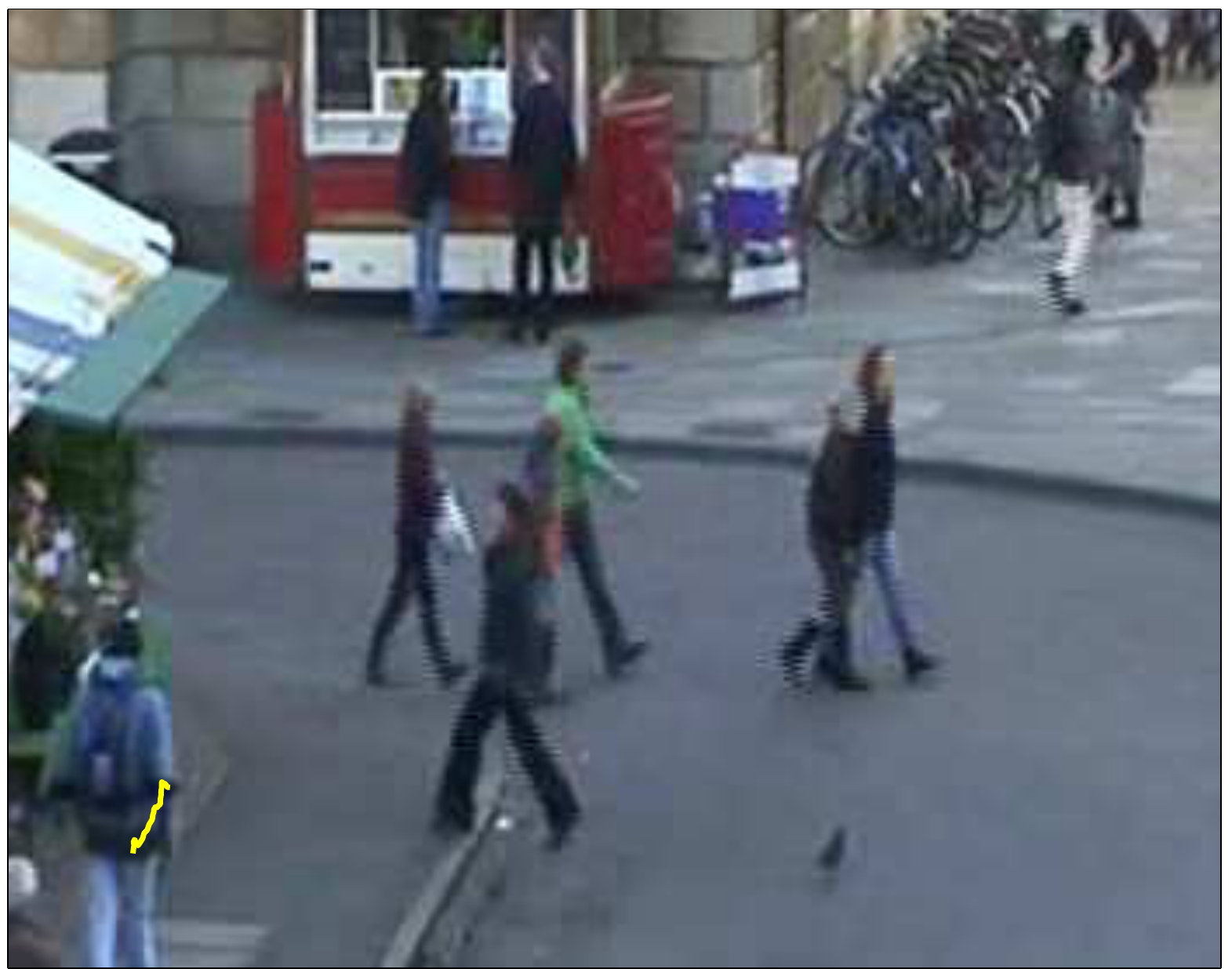}&
    \includegraphics[width=0.45\textwidth]{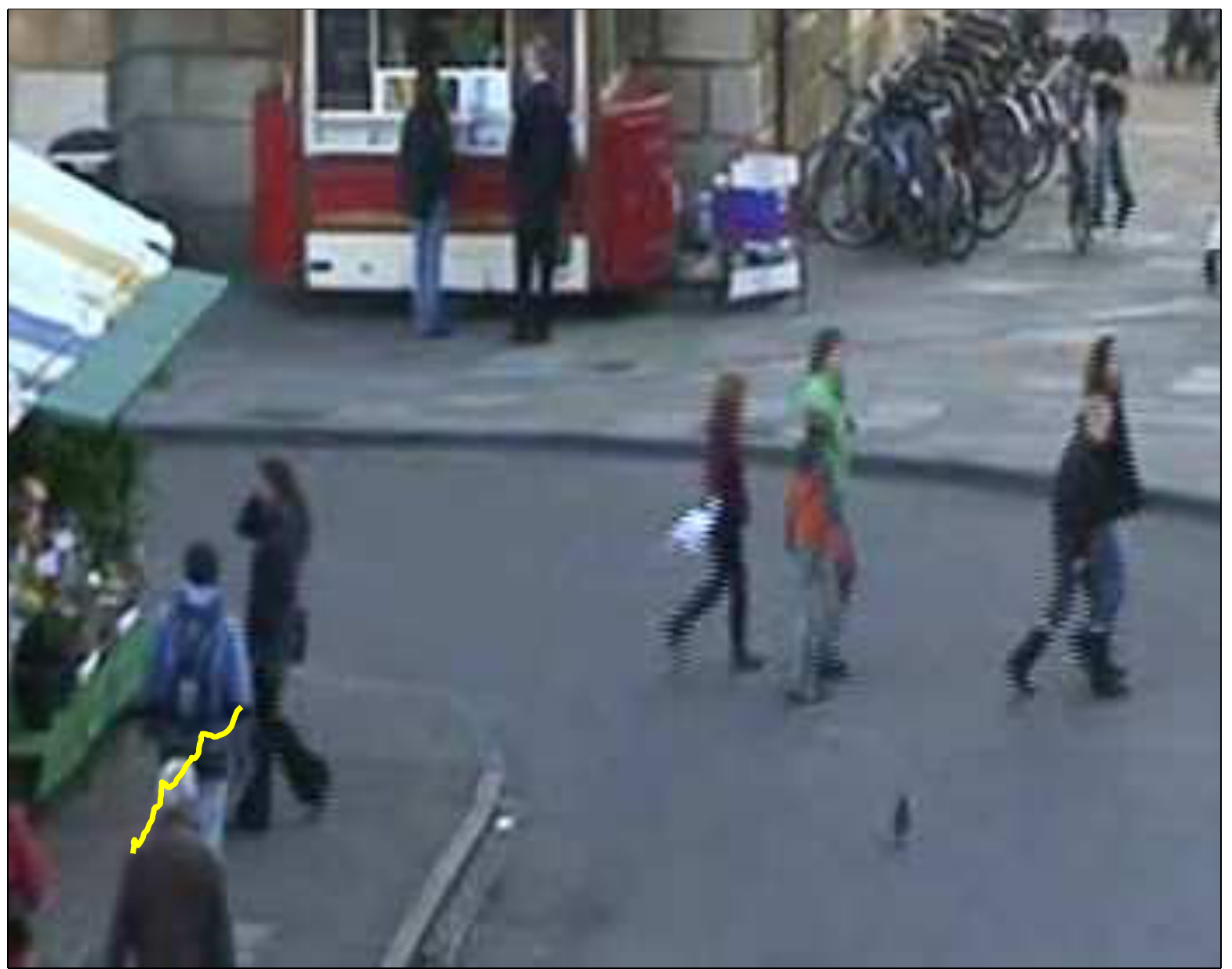}\\
    Feature discovery & Successful tracking \\&\\
    \includegraphics[width=0.45\textwidth]{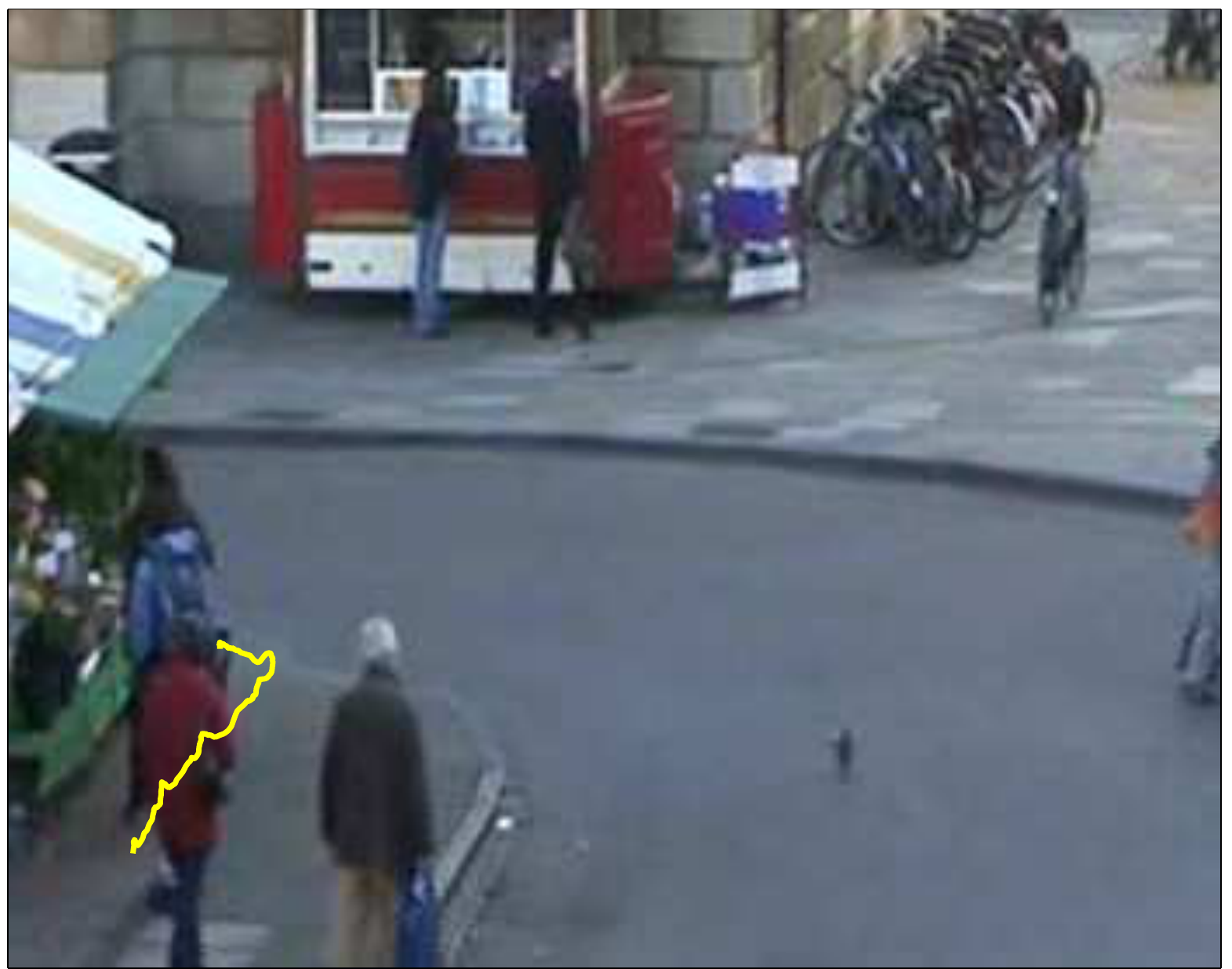}&
    \includegraphics[width=0.45\textwidth]{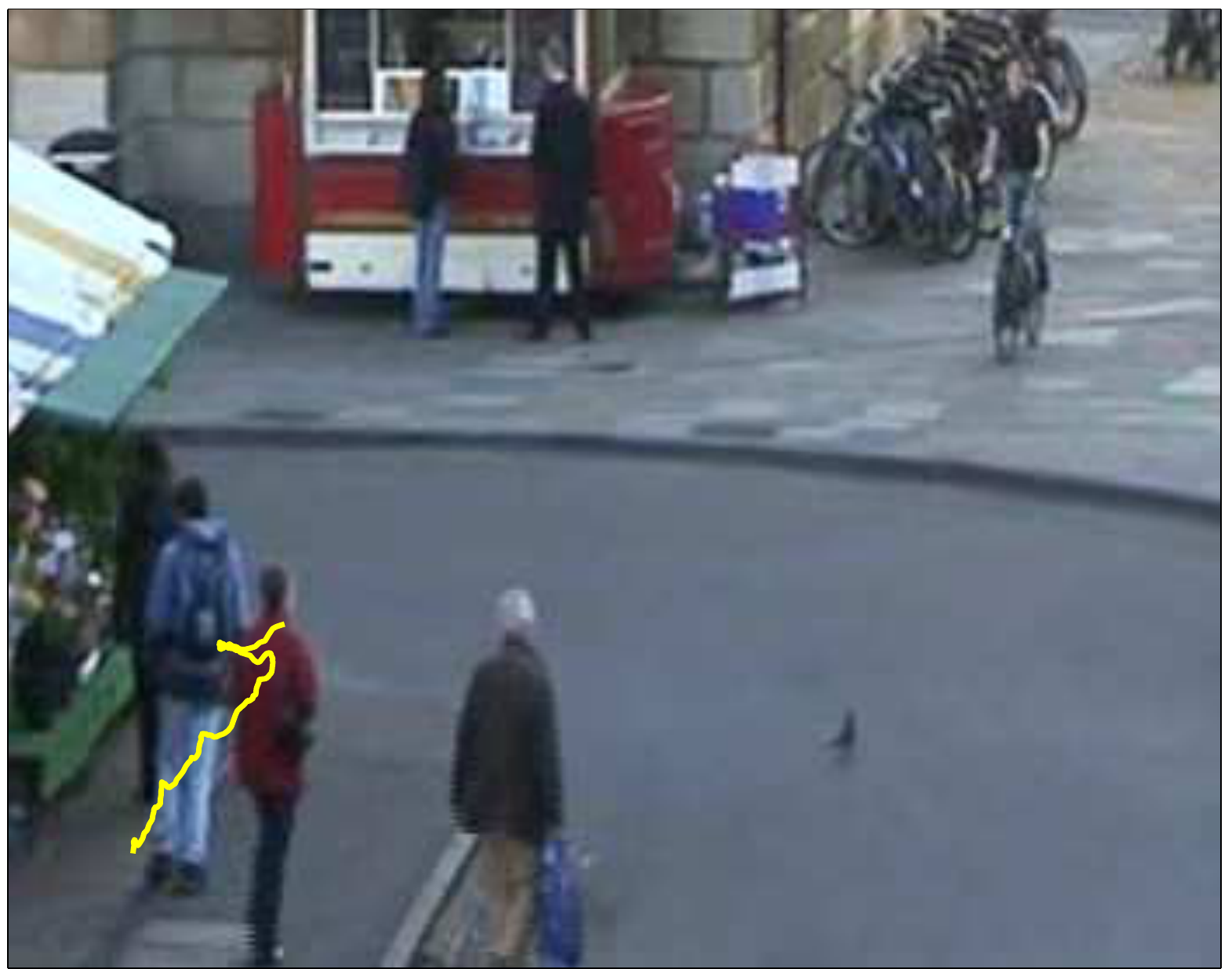}\\
    Successful tracking & Incorrect feature matching\\&\\
    \includegraphics[width=0.45\textwidth]{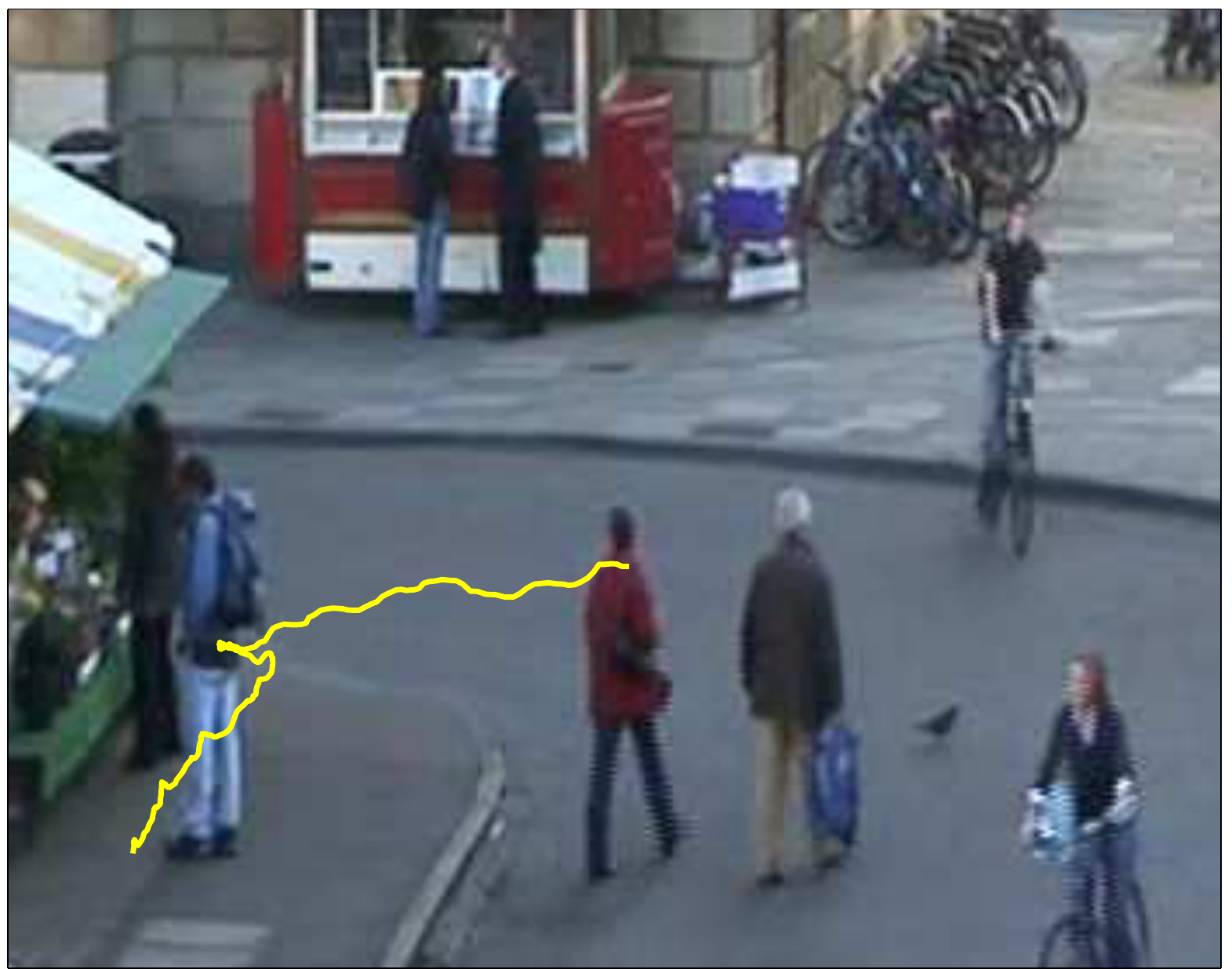}&
    \includegraphics[width=0.45\textwidth]{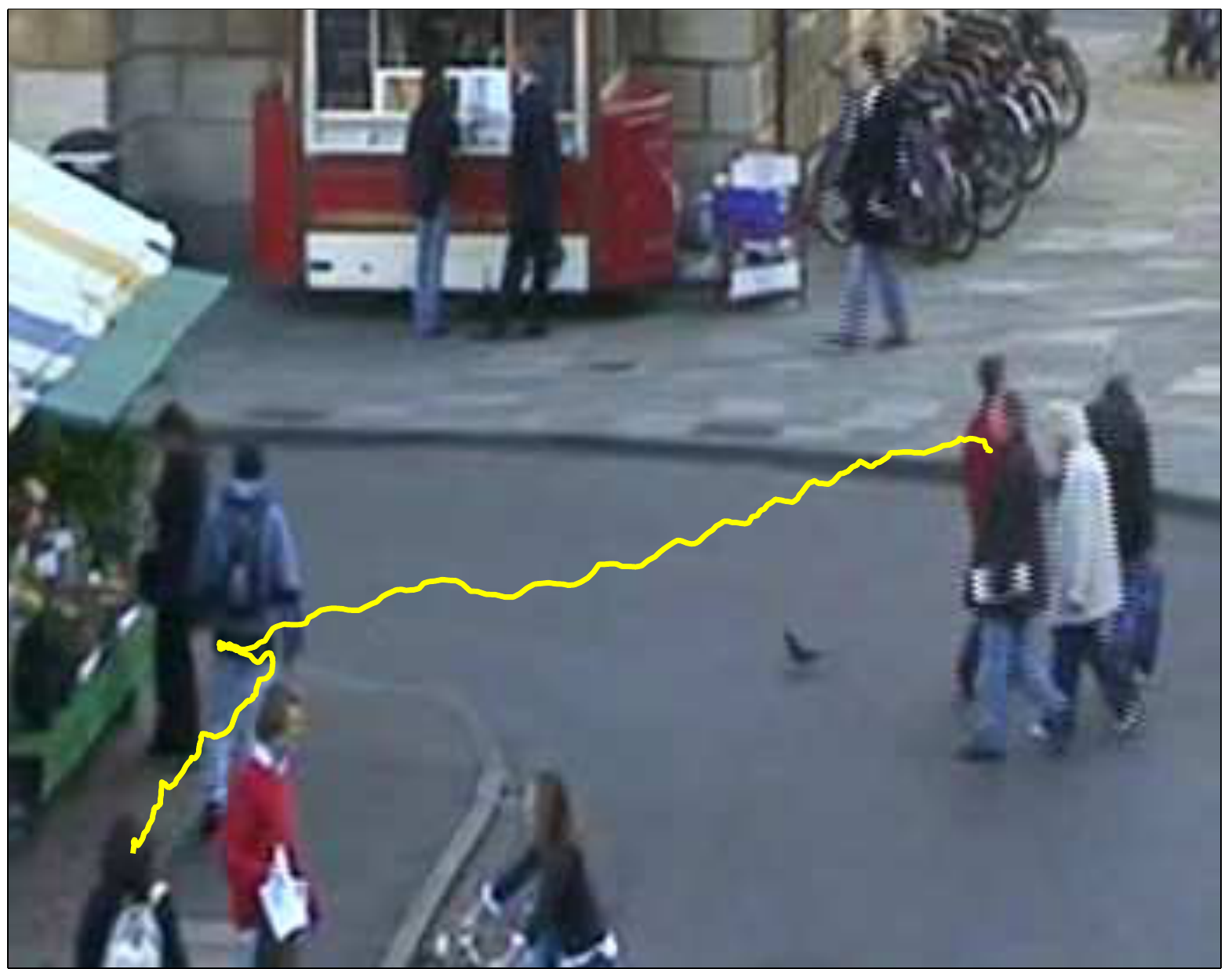}\\
    Incorrect feature is tracked & Incorrect feature is tracked\\&\\
  \end{tabular*}
  \caption{ The principal source of errors of our algorithm arises not at the higher level of processing,
            concerned with how motion is modelled, but rather at the lowest level of motion extraction.
            Reliable many body tracking in crowded scenes is an outstanding problem.
            }
  \label{f:err}
\end{figure}

\section{Summary and conclusions}
In this paper we addressed the problem of unusual behaviour detection in busy public places by means of novelty detection in holistic body motion. Our general approach is bottom-up. We argued that in the scenario of interest scene motion is best extracted using local features. The obtained feature tracks are segmented into tracklets, which are used to infer scene-specific tracklet primitives using a custom clustering algorithm. Finally, we described two motion models based on tracklet primitives: a multi-scale but non-hierarchial ensemble of Markov chains and an alternative which takes into account human motion planning by learning the distributions of paths lengths between tracklet primitives.

\bibliographystyle{plain}
\bibliography{my_bibliography}

\end{document}